\documentclass[runningheads]{llncs}
\usepackage{eccv}
\usepackage{eccvabbrv}
\usepackage{graphicx}
\usepackage{booktabs}
\usepackage{algorithm}
\usepackage{algpseudocode}
\usepackage{titletoc}
\usepackage{wrapfig}
\usepackage{pifont}
\usepackage[accsupp]{axessibility}  

\usepackage{hyperref}

\usepackage{orcidlink}
\definecolor{gray}{rgb}{0.5,0.5,0.5}
\usepackage{color}
\usepackage{xcolor}
\usepackage{colortbl,booktabs}

\newcommand{\proposed}{\textbf{\texttt{Mew}}}

\begin{document}

\title{\includegraphics[height=1cm]{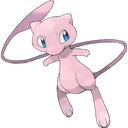} \texttt{\textbf{Mew}}: Multiplexed Immunofluorescence Image Analysis through an Efficient Multiplex Network}

\titlerunning{\proposed}

\author{Sukwon Yun\inst{1}\orcidlink{0000-0002-5186-6563},
Jie Peng\inst{2}\orcidlink{0000-0003-3805-9326},
Alexandro E. Trevino\inst{3}\orcidlink{0000-0002-9763-1540}, \\
Chanyoung Park\inst{4}\orcidlink{0000-0002-5957-5816},
Tianlong Chen \inst{1,5,6}\orcidlink{0000-0001-7774-8197}}

\authorrunning{S. Yun et al.}

\institute{
University of North Carolina at Chapel Hill
\email{\{swyun,tianlong\}@cs.unc.edu} \and
University of Science and Technology of China 
\email{pengjieb@mail.ustc.edu.cn} \and
Enable Medicine
\email{alex@enablemedicine.com} \and
KAIST
\email{cy.park@kaist.ac.kr} 
\and MIT 
\and Harvard University
}

\maketitle

\vspace{-5mm}

\begin{abstract}
  Recent advancements in graph-based approaches for multiplexed immunofluorescence (mIF) images have significantly propelled the field forward, offering deeper insights into patient-level phenotyping. However, current graph-based methodologies encounter two primary challenges:~\ding{172}~\underline{Cellular Heterogeneity}, where existing approaches fail to adequately address the inductive biases inherent in graphs, particularly the homophily characteristic observed in cellular connectivity; and ~\ding{173}~\underline{Scalability}, where handling cellular graphs from high-dimensional images faces difficulties in managing a high number of cells. To overcome these limitations, we introduce~\proposed, a novel framework designed to efficiently process \underline{\textbf{\texttt{m}}}IF images through the lens of multipl\underline{\textbf{\texttt{e}}}x net\underline{\textbf{\texttt{w}}}ork.~\proposed~innovatively constructs a multiplex network comprising two distinct layers: a Voronoi network for geometric information and a Cell-type network for capturing cell-wise homogeneity. This framework equips a scalable and efficient Graph Neural Network (GNN), capable of processing the entire graph during training. Furthermore,~\proposed~integrates an interpretable attention module that autonomously identifies relevant layers for image classification. Extensive experiments on a real-world patient dataset from various institutions highlight~\proposed's remarkable efficacy and efficiency, marking a significant advancement in mIF image 
  analysis. The source code of~\proposed~can be found here: \url{https://github.com/UNITES-Lab/Mew}

  
\end{abstract}

\section{Introduction}
\label{sec:intro}

\begin{wrapfigure}[6]{r}{0.22\textwidth}
    \centering
    \includegraphics[width=0.8\linewidth]{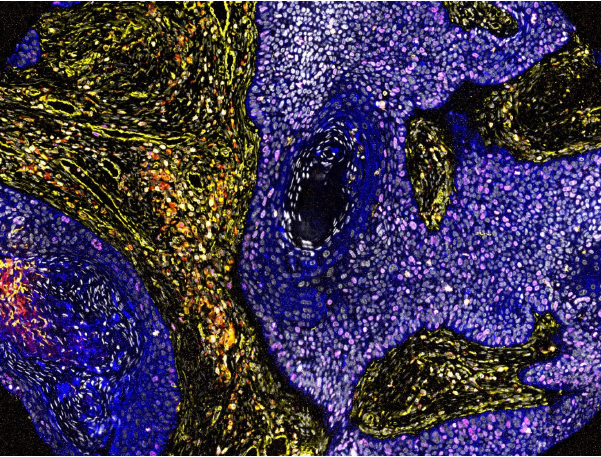}
    \caption{mIF image.}
    \label{fig:mif_example}
\end{wrapfigure}

Multiplexed immunofluorescence (mIF) imaging is a pivotal technique for the simultaneous detection and visualization of multiple protein targets within a single tissue sample. Utilizing antibodies labeled with diverse fluorescent dyes, mIF enables the identification and quantification of numerous biomarkers at the cellular level.  This method offers a comprehensive view of the cellular composition and spatial interplay within tissue microenvironments, proving invaluable in oncology, immunology, and pathology~\cite{mIF1, mIF2, mIF3, macro_tumor, macro_tcell}, as shown in Figure~\ref{fig:mif_example}.

\noindent Recently, the adoption of graph-based ML in the analysis of mIF images, has shown promising results. This approach effectively incorporates geometric data about cellular composition in tissue samples. A notable example is SPACE-GM~\cite{space-gm}, which generates graphs from CODEX using Voronoi diagrams derived from Delaunay triangulation. By treating tumor microenvironments as localized subgraphs, SPACE-GM employs GNNs~\cite{gin} to capture complex cellular interactions, thereby enabling the prediction of differential clinical outcomes. While leveraging GNNs to capture local semantics appears rational, two significant foundational bottlenecks remain:

\begin{figure}[!t]
    \centering
    \includegraphics[width=1\columnwidth]{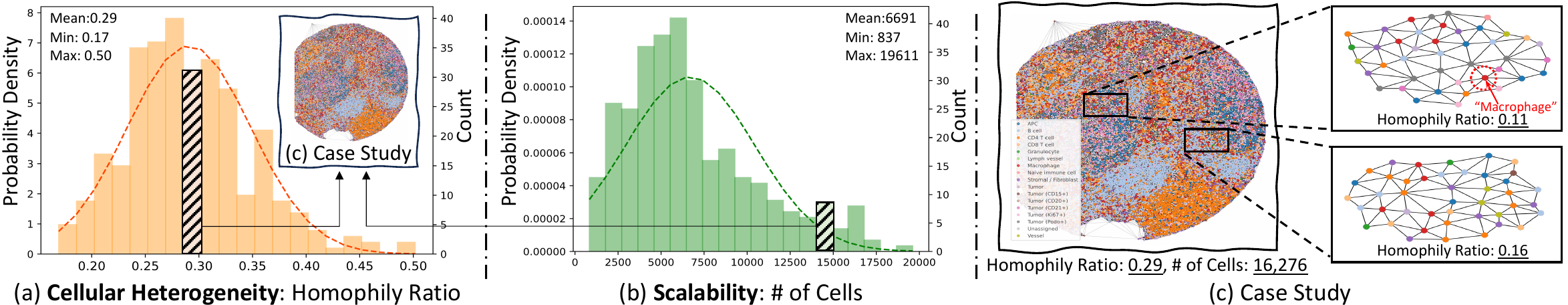}
    \caption{Distribution of mIF image is examined through (a) \textbf{Cellular Heterogeneity}: Homophily Ratio and (b) \textbf{Scalability}: number of cells with (c) case study of mIF cellular graphs in the UPMC dataset. Homophily Ratio: $\frac{\# \text{of edges connecting same cell-type nodes}}{\# \text{of total edges}}$. Distributions on other datasets are provided in Appendix~\ref{appendix:A}.}
    \label{fig:motivation}
\end{figure}

\noindent{\underline{{\textbf{\ding{172} Cellular Heterogeneity.}}} When leveraging GNNs, it is crucial to consider the fundamental inductive bias of GNNs, namely, \textit{the homophily assumption: the birds of a feature flock together}~\cite{gcn, gat, graphsage, birds}. This assumption posits that neighboring nodes typically belong to the same class, thereby possessing similar representation with their neighbors. However, as depicted in Figure~\ref{fig:motivation} (a), we observe that the homophily ratio of mIF cellular graphs predominantly ranges between 0.25 and 0.30. This scenario in the mIF domain starkly contrasts with the typical inductive bias of GNNs, where the homophily ratio often surpasses 0.7, as seen in citation networks~\cite{cora, citation}. Analyzing graphs in such a heterogeneous environment, i.e., accurately phenotyping patients, presents significant challenges. For instance, Figure~\ref{fig:motivation} (c) reveals the existence of diverse heterogeneous connections, such as those between `Macrophage' cells and tumor-related cells or CD T immune cells. Focusing solely on local heterogeneous connections might obscure the accurate determination of `Macrophage' cells' role, particularly relevant to patient-level phenotyping. As a remedy, highlighting the self-connections of `Macrophage' cells in a broader context is essential.  By achieving a homophily ratio of 1.0, this approach provides essential insights for effectively managing cellular heterogeneity, offering a strategic advantage in navigating and addressing the complexities of the tumor environment and patient-level phenotyping. Biologically, supported by literature~\cite{macro_homo1, macro_homo2}, analyzing interactions with neighboring cells of similar roles throughout the entire image discloses whether macrophages mainly exhibit tumor-suppressive or tumor-promoting behaviors. In conclusion, our findings underscore the importance of integrating self-connections among identical cell-types across the entire image to fully leverage the GNN framework in mIF image analysis, thus highlighting GNNs' core inductive bias.

\noindent{\underline{{\textbf{\ding{173} Scalability.}}} Moreover, a significant challenge in mIF analysis is scalability, particularly in handling the vast volume of image data encompassing a high number of cells. As depicted in Figure~\ref{fig:motivation} (b) and (c), transforming each image into a graph at the cellular level results in high complexity due to a large number of nodes, setting the mIF image domain apart from standard graph-based ML tasks, which typically focus on analyzing a single graph. To tackle this issue, SPACE-GM introduces a 3-hop neighbor sampling strategy. However, this approach, by relying on a localized subset of nodes, limits the model's capacity to grasp global information—vital for patient-level phenotyping—and restricts GNN's parameter updates to a limited scope. Moreover, the feasibility of storing 3-hop neighbor graph data in chunks within computational resources for training efficiency poses practical concerns, especially in clinical settings where memory resources are limited and prompt predictions are essential for patient care. This scenario underscores the necessity for developing a method capable of handling large graph sizes in a scalable and resource-efficient manner.  

\

\noindent{{{\textbf{Our apporach.}}} Building upon these motivations (\ding{172}+\ding{173}), we introduce a novel framework for \underline{\textbf{\texttt{m}}}IF image analysis via multipl\underline{\textbf{\texttt{e}}}x net\underline{\textbf{\texttt{w}}}ork, called~\proposed\footnote{The terms 'multiplexed' and 'multiplex' are used distinctly in this paper: 'multiplexed' refers to image data, while 'multiplex' refers to a network type. Despite this distinction, both terms derive from a root meaning 'multi-layered'.}. The core idea of~\proposed~is to tackle both the geometric information and cell-type heterogeneity in mIF images by employing an efficient multiplex network framework for enhanced analysis. For each mIF image, we create a Voronoi network based on cell coordinates and simultaneously establish a cell-type network, connecting nodes of identical cell types. This dual-network setup is then processed by a scalable Graph Neural Network, which employs a precomputing technique for message-passing operations, thus boosting scalability and efficiency during training. Moreover,~\proposed~equips the Voronoi-Cell-type Attention module, which skillfully determines the relevance of each network layer in concluding the patient's phenotype. Extensive testing on real-world patient datasets from three distinct institutions has thoroughly validated the efficacy and efficiency of~\proposed. Our three principal contributions are summarized as follows:

\begin{itemize}
\item [$\star$] We pioneer the exploration of homophily characteristics of cell types in mIF image analysis within a multiplex network framework, marking the first application of a multiplex network's in the mIF image domain to our knowledge.
\item [$\star$]~\proposed~introduces a scalable Graph Neural Network architecture featuring a novel stochastic edge sampling technique that eliminates the necessity of saving data chunks, thereby efficiently utilizing the entire graph during training for improved efficiency and scalability.
\item [$\star$] Our comprehensive experimental evaluations on real-world mIF datasets (including datasets from UPMC, Stanford, and DFCI) consistently affirm the robustness and superior performance of~\proposed.
\end{itemize}

\section{Related Works}
\paragraph{Multiplexed Immunofluorescence Image Analysis.}
Multiplexed immunofluorescence (mIF) imaging is a sophisticated technique that allows for the simultaneous detection and quantification of multiple biomarkers within a single tissue section, providing a comprehensive view of the cellular and molecular landscapes. The field of mIF image analysis has witnessed considerable evolution, transitioning from early methods employing traditional image processing techniques~\cite{traidtional_image0,traidtional_image1,traidtional_image2,traidtional_image3,mIF1,mIF2,morpho_image,hu2021e2vts,hu20212020} to the latest advancements that integrate spatial information with deep learning~\cite{prml,deeplearning_ian,unet,image_to_image,transformer,efficient_net,mIF3,mIF4,mIF5,spatial1,spatial2,yu2022unified,shen2021umec}. Initially, mIF approaches predominantly relied on spectral unmixing and manual annotation for identifying and quantifying cellular markers. Introducing convolutional neural networks (CNNs) marked a significant leap forward, enhancing cell classification, feature extraction, and biomarker detection. Notably, the adoption of deep learning architectures like U-Net~\cite{unet} and 3D U-Net~\cite{3d_unet} for dense volumetric segmentation has set new benchmarks in the field. Moreover, the application of GANs for data augmentation~\cite{gan, dcgan, cgan, gan_appication_1, gan_appication_2, shen2021learning} and the innovative use of transformers~\cite{transformer} for capturing long-range dependencies within images have further enriched the analytical capabilities in mIF analysis. Recent research frontiers have moved towards incorporating spatial information, with innovations like SPACE-GM~\cite{space-gm} exemplifying this trend. SPACE-GM, leveraging spatial cellular graphs, employs graph-based ML models to elucidate intricate spatial cell interactions, thereby enriching our understanding of the tissue microenvironment. Despite these advancements, optimally addressing cellular heterogeneity in the context of graph-based ML remains a critical challenge to be addressed.

\paragraph{Graph Neural Networks.} Graph-based ML has seen a surge of innovation with the development of various models, each marking a milestone in the field. Graph Convolutional Network (GCN)\cite{gcn} played a crucial role in popularizing Graph Neural Networks (GNNs) by adapting convolutional principles to graphs. Building upon this, the Graph Attention Network (GAT)\cite{gat, gatv2} introduced attention mechanisms, enabling refined weighting of node interactions. GraphSAGE~\cite{graphsage} furthered this progression by facilitating inductive learning via a novel neighborhood sampling and aggregation approach. Addressing scalability, the Scalable Inception Graph Neural Networks (SIGN)\cite{sign} streamlined the application to larger graphs through precomputed neighborhood information. ClusterGCN\cite{clustergcn} tackled large-scale graph learning by employing graph clustering techniques. In the realm of heterogeneous graphs, Deep Multiplex Graph Infomax (DMGI)\cite{dmgi} leveraged unsupervised learning across varied node and edge types. More recently, High-order Deep Multiplex Infomax (HDMI)\cite{hdmi} has innovatively incorporated high-order mutual information for self-supervised node embedding. Collectively, these models have significantly advanced graph-based ML, addressing challenges like scalability, heterogeneity, and unsupervised learning. Despite these advancements, their application to extensive spatial image datasets, such as mIF data, remains underexplored, particularly in terms of cellular heterogeneity and scalability.

\begin{figure*}[!t]
    \centering
    \includegraphics[width=\linewidth]{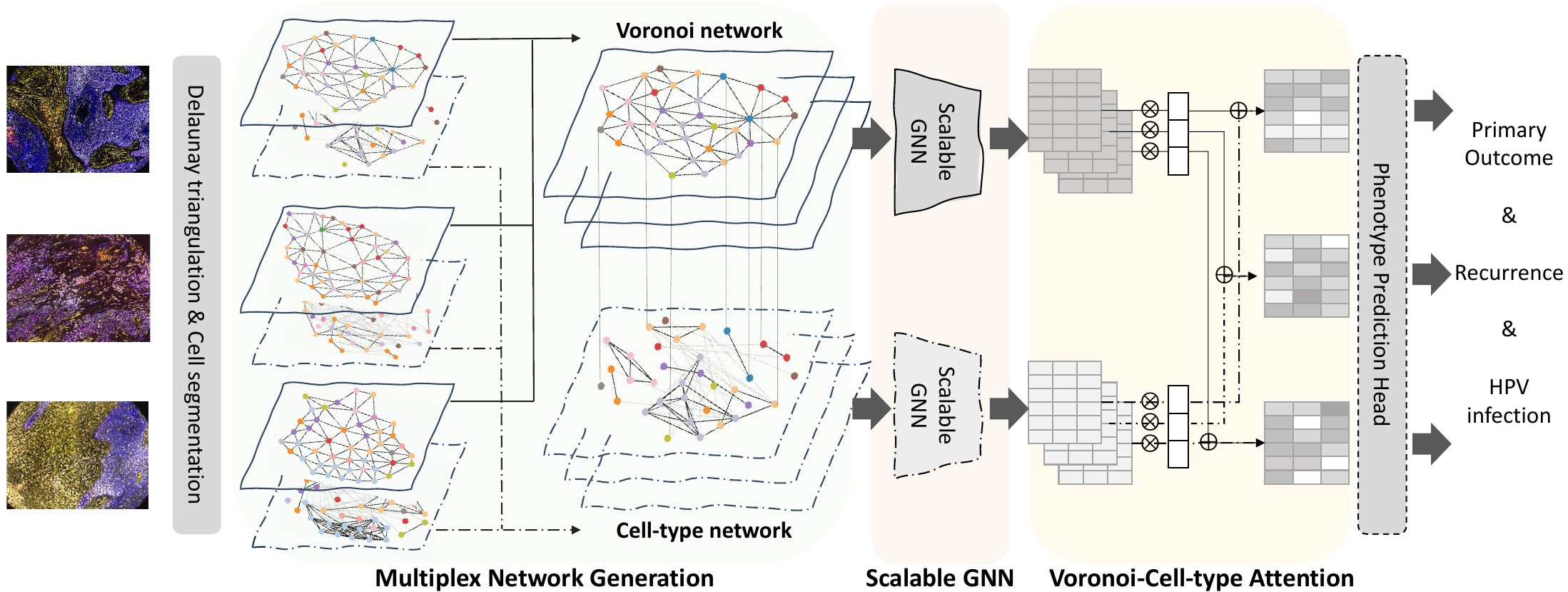}
    \caption{Overall framework of~\includegraphics[height=0.5cm]{figs/mew2.png}~\proposed. Given mIF images, it employs Delaunay triangulation and cell segmentation, leading to the formation of a multiplex network composed of two distinct networks: a Voronoi network and a Cell-type network. Here, two networks share common nodes but have distinctive edge connections. These networks are then analyzed using scalable GNNs equipped with precomputing capabilities and stochastic edge sampling techniques. This analysis is further enhanced by the Voronoi-Cell-type Attention, a mechanism designed for evaluating the significance of each layer. Ultimately, it predicts the patient's phenotype via the phenotype prediction head.}
    \label{fig:main_figure}
\end{figure*}

\section{Methodology}
In this section, we present~\proposed, a novel algorithm tailored for mIF image analysis through a multiplex network approach, utilizing Voronoi and Cell-type networks. Initially, for a given mIF image, preprocessing steps like Delaunay triangulation and cell segmentation are executed to derive two distinct networks based on different edge types to generate a multiplex network (Sec 3.1). Keeping the multiplex network concept central, we apply scalable Graph Neural Networks for efficient training in a whole graph perspective with a stochastic edge sampling technique (Sec 3.2). This is followed by the Voronoi-Cell-type Attention, which seamlessly integrates the information of each network (Sec 3.3), culminating in the prediction of the patient's phenotype. Figure~\ref{fig:main_figure} illustrates an overview of~\proposed~and the comprehensive training algorithm is provided at Appendix~\ref{appendix:B}.

\paragraph{Task - Patient-level Phenotyping: Binary Classification and Hazard Modeling.} For each image $I$ in a set of patient samples with a total number of $I_{N}$, we transform these images into a series of graphs $\mathcal{G}^1, \mathcal{G}^2, \ldots, \mathcal{G}^{I_{N}}$. Each graph, $\mathcal{G}^i$, for all $i \leq I_{N}$, consists of a set of nodes and edges that represent sample $i$ and its associated phenotypes $\mathcal{Y}^i$. These phenotypes are categorized into two groups: Binary Classification and Hazard Modeling. For each phenotype prediction task, a multi-task learning approach is adopted, which integrates the losses from each task, such as the primary outcome or recurrence in the case of binary classification. Reformulated as a graph classification task within a supervised learning framework, the primary objective of~\proposed~is to develop a scalable, efficient graph model that can accurately predict the phenotype of a given sample.

\subsection{Multiplex Network Generation}
This section outlines the process of generating a multiplex network from spatial cellular graphs, incorporating two distinct relational types. Initially, with each sample provided as an image, constructing a graph structure becomes imperative to harness the geometry information. This involves leveraging a message-passing scheme with connected neighbors. To extract geometric data like cell coordinates, we utilize the preprocessing approach of SPACE-GM~\cite{space-gm}. Here, cell nuclei, identified using the DeepCell neural-network segmentation tool~\cite{deepcell}, are processed with segmentation masks to yield 2D cellular coordinates represented as discrete points in Euclidean space. Subsequently, to delineate neighborhood relations, Delaunay triangulation was applied to these cellular centroids. This results in Voronoi diagrams linked via circumcircle centers. Each image is thus transformed into a graph where nodes represent cellular centroids and edges denote connections between neighboring Voronoi polygons. Formally, this graph is represented as follows:

\begin{equation}
\mathcal{G}^{I} = (\mathcal{V}^{I}, \mathcal{E}^{I}), \forall I \in \{1, \ldots, I_{N}\}
\end{equation}

\noindent where \(\mathcal{G}^{I}\) represents the Voronoi graph of image sample \(I\), comprising a set of cellular nodes \(\mathcal{V}^{I}\) and their connecting edges \(\mathcal{E}^{I}\).

Concurrently, individual cell biomarkers identified by CODEX are processed through principal component analysis (PCA) and Louvain graph clustering~\cite{louvain}, yielding cell-type annotations for each cell. Diverging from existing approaches that use cell-type as a mere feature within a concatenated matrix, we adopt an orthogonal strategy. We argue that cell-type information can provide distinctive direction besides the geometry information, which contains shared expressions of biomarkers, and overall cell-type populations can play a crucial role in graph classification tasks like predicting primary outcomes or HPV infection. This approach addresses the cellular heterogeneity observed in Figure~\ref{fig:motivation} (a) through message-passing exclusively among nodes sharing the same cell-type. Formally, for a set of cell-types \(\mathcal{C}^{I}\) in image \(I\), they are depicted as follows:
\begin{equation}
\begin{gathered}
\mathcal{G'}^{I} = (\mathcal{V'}^{I}, \mathcal{E'}^{I}), \; \forall I \in \{1, \ldots, I_{N}\} \\
\text{where} \; \mathcal{E'}^{I} = \{(\mathcal{V'}^{I}_{i}, \mathcal{V'}^{I}_{j}) \mid \mathcal{C}_i^{I} = \mathcal{C}_j^{I}, \forall i, j \in \{1, ..., |\mathcal{V'}^{I}|\}, i \neq j \}
\end{gathered}
\end{equation}

\noindent where $\mathcal{G'}^{I}$ represents the additional graph composed of nodes $\mathcal{V'}^{I}$ and edges $\mathcal{E'}^{I}$ that connect nodes sharing the same cell-type. It's crucial to note that, since the nodes (i.e., cell nuclei) originate from the same image $I$, the node sets are identical ($\mathcal{V}^{I}=\mathcal{V'}^{I}$), as depicted in Figure~\ref{fig:main_figure}. However, the relationship type varies ($\mathcal{E}^{I}\neq\mathcal{E'}^{I}$). Incorporating this cell-type network results in a multiplex (multi-layered) network as follows:

\begin{equation}
\tilde{\mathcal{G}}^{I} = (\mathcal{G}^{I}, \mathcal{G'}^{I})
\end{equation}

\noindent where $\tilde{\mathcal{G}}^{I}$ denotes a multiplex network comprising the Voronoi network $\mathcal{G}^{I}$, which captures local geometric information, and the cell-type network $\mathcal{G'}^{I}$, responsible for representing cell type composition across the entire image. The incorporation of the cell-type layer is expected to enhance the representation of cells sharing the same cell type, thereby optimizing the message-passing scheme from a global perspective. This approach aligns with the fundamental inductive bias of GNNs, specifically the homophily assumption, by ensuring that cells of similar types are more effectively connected and represented within the network. 

\subsection{Scalable Graph Neural Network}

\begin{wrapfigure}[16]{r}{0.36\textwidth}
\centering
\includegraphics[width=0.9\linewidth]{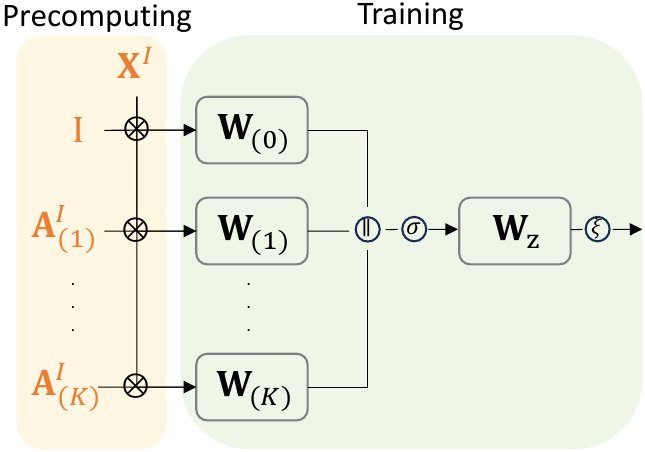}
\caption{Illustration of Precomputing and Training procedure. Once the orange colored components, \({\mathbf{X}^{I}}\), \({\mathbf{A}^{I}_{(1)}\mathbf{X}^{1}}\), $\cdots$, \({\mathbf{A}^{I}_{(K)}\mathbf{X}^{I}}\) are precomputed, they are utilized throughout the training and inference phases.}
\label{fig:precomputing}
\end{wrapfigure}

Now, given a multiplex network, the challenge lies in efficiently handling the graph input to derive node embeddings rich in information relevant to phenotype prediction. Direct application of Graph Neural Networks (GNN)\cite{chebynet, gcn, gat} to the dataset is hindered by scalability challenges, as detailed in Figure~\ref{fig:motivation} (b). The number of cells per sample in the dataset varies widely, from 837 to 19,611, with an average of 6,691 cells, culminating in a total of 2,061,066 cells. This presents a significant challenge compared to current Graph Neural Networks, typically benchmarked on datasets~\cite{citation, coauthor} with a single graph of 2,000-19,000 nodes. 
To address such scalability issues, inspired by the efficient precomputation and fast training and inference of SIGN~\cite{sign}, we apply a similar precomputing strategy first to the Voronoi network ($\mathcal{G}^{I}$), which can be described as follows:

\begin{equation}
\label{eqn:sign}
\begin{gathered}
\mathbf{H}^{I} = \sigma{([{\mathbf{X}^{I}}\mathbf{W}_{(0)}, {\mathbf{A}^{I}_{(1)}\mathbf{X}^{I}} \mathbf{W}_{(1)}, \ldots, {\mathbf{A}^{I}_{(K)}\mathbf{X}^{I}}\mathbf{W}_{(K)}])} \\
\mathbf{Z} = \xi({\mathbf{H}^{I}\mathbf{W}_{z}})
\end{gathered}
\end{equation}

\noindent where \(\mathbf{H}^{I} \in \mathbb{R}^{|\mathcal{V}^{I}|\times D(K+1)}\) is a concatenated embedding matrix for \(|\mathcal{V}^{I}|\) nodes in image $I$, encompassing the original matrix and $K$-hop elements, each transformed into a hidden dimension \(D\). \(\mathbf{A}^{I}_{(k)}\in \mathbb{R}^{|\mathcal{V}^{I}|\times |\mathcal{V}^{I}|}\), $\forall k \leq K$ represents the symmetrically normalized adjacency matrix (\(\mathbf{D}^{-1/2 }\hat{\mathbf{A}}^{I}\mathbf{D}^{-1/2}\), with self-loops added, \(\hat{\mathbf{A}}\), and its corresponding degree matrix \(\mathbf{D}\)) after \(k\) iterations of multiplication, capturing up to \(k\)-hop neighbor cells in image \(I\). \(\mathbf{W}_{(k)} \in \mathbb{R}^{F\times D}\) is a weight matrix transforming the original feature space \(F\) into the embedding space \(D\) for each \(k\). The concatenated output, comprising message-aggregated (\(\mathbf{A}\mathbf{X}\)) matrices and parameters, now generates the embedding matrix\footnote{We use an output embedding matrix instead of a predicted class matrix to align with the subsequent attention module, leveraging the embedding representation.}, \(\mathbf{Z} \in \mathbb{R}^{|\mathcal{V}^{I}|\times D}\), processed by the final weight matrix \(\mathbf{W}_{z} \in \mathbb{R}^{D(K+1) \times D}\) with nonlinearities \(\sigma\) and \(\xi\). Notably, these elements \({\mathbf{X}^{I}}\), \({\mathbf{A}^{I}_{(1)}\mathbf{X}^{1}}\), \({\mathbf{A}^{I}_{(k)}\mathbf{X}^{I}}\) can be efficiently precomputed and obtained via techniques like Apache Spark~\cite{apache} or sparse matrix multiplication~\cite{pytorch} before entering training, as they do not depend on learnable model parameters and remain static during training. This approach significantly enhances our algorithm's efficiency, which will be discussed later in Sec~\ref{sec:scalability}. By performing just $K$ matrix multiplications per image, we efficiently generate the requisite input for the feedforward network, circumventing the need for repetitive multiplications with the adjacency matrix. This efficient method substantially conserves computational resources and reduces processing time. In contrast, conventional GNNs necessitate recalculating the message-aggregation process \(\mathbf{A}\mathbf{X}\) in every iteration, incorporating nonlinear transformations through continually updated learnable weight matrices. A visual depiction of this procedure is provided in Figure~\ref{fig:precomputing}, illustrating the streamlined computational flow.

\paragraph{Discussion on Message-Aggregation in the Cell-type Network.} Recall that we are working with a multiplex network ($\tilde{\mathcal{G}}$), where the cell-type network ($\mathcal{G}^{'}$) plays a pivotal role alongside the Voronoi network ($\mathcal{G}$). One might consider employing scalable GNNs as discussed in Equation~\ref{eqn:sign}. However, since all nodes of the same cell type are directly connected in $\mathcal{G}^{'}$, each node accesses its neighbors in just 1-hop, as shown in Figure~\ref{fig:main_figure} (refer to the Cell-type network). Although effective for aggregating information from the same cell-type nodes, this 1-hop approach does not incorporate new information beyond the 2-hop range, limiting the message-passing scheme’s generalizability, particularly in distinguishing between short-range and long-range connections. 

To address this and enhance generalizability, we introduce a new \textbf{Stochastic Edge Sampling} technique that samples edge indices based on their distance. A simple method is to introduce an edge-deleting hyperparameter and delete edges ($\mathcal{E}^{'}$) in the cell-type network. However, selecting a fixed hyperparameter could disrupt meaningful connections by not adequately considering the semantics of neighboring and distant relationships. Our method capitalizes on the normalized distances between cellular centroids derived from Voronoi polygons. Contrary to SPACE-GM, which considers edges shorter than 20 \textmu m as neighboring and designs separate embeddings based on edge length, we directly use distance as a probability for edge sampling. These distances, normalized between 0 and 1, role as the main resource for our stochastic edge sampling strategy. We construct a distance pair matrix, $\mathbf{P}^{I} \in \mathbb{R}^{|\mathcal{V}^{I}|\times|\mathcal{V}^{I}|}$, and its complement, $\mathbf{\hat{P}}^{I} = \mathbf{1} - \mathbf{P}^{I}$, adhering to the principle that closer nodes are more influential~\cite{birds}. Biologically, the importance of proximity among identical cell types is underscored by studies~\cite{neighbor1, neighbor2} indicating that neighboring cells of the same type often form local clusters or specialized microenvironments. 

In essence, our stochastic edge sampling method aims to differentiate the influence of neighboring versus distant nodes, thereby enriching the model's ability to capture biologically significant interactions. This approach can be formally expressed as follows:

\begin{equation}
\label{eqn:stochastic}
\mathbf{A'}^{I}_{ij} = \begin{cases}
    1 & \text{if}\;\; \text{Bernoulli}(\mathbf{\hat{P}}^{I}_{ij}) = 1 \\
    0 & \text{Otherwise}  
\end{cases}
\end{equation}

\noindent where $\mathbf{A'}^{I}$ represents the newly sampled adjacency matrix for the cell-type network. This operation is conducted for each hop ($k$), utilizing the sampled adjacency matrix as the basis for message-aggregation in the cell-type network, which follows as below:

\begin{equation}
\label{eqn:sign_cell}
\begin{gathered}
\mathbf{H'}^{I} = \sigma{([{\mathbf{X}^{I}}\mathbf{W'}_{(0)}, {\mathbf{A'}^{I}_{(1)}\mathbf{X}^{I}} \mathbf{W'}_{(1)}, \ldots, {\mathbf{A'}^{I}_{(K)}\mathbf{X}^{I}}\mathbf{W'}_{(K)}])} \\
\mathbf{Z'} = \xi({\mathbf{H'}^{I}\mathbf{W'}_{z}})
\end{gathered}
\end{equation}

\noindent where $\mathbf{H'}^{I}$ represents the concatenated embedding matrix of precomputed inputs and their subsequent feedforward transformation. Here, \(\mathbf{W'}_{(k)} \in \mathbb{R}^{F\times D}\) is a weight matrix\footnote{Given the multiplex network nature, the weight matrix can either be shared with the Voronoi network ($\mathbf{W'}_{(i)} = \mathbf{W}_{(i)}$) or kept separate ($\mathbf{W'}_{(i)} \neq \mathbf{W}_{(i)}$). For flexibility, we treat the use of a shared weight matrix as a hyperparameter.} for cell-type network transforming the original feature space \(F\) into the embedding space \(D\) for each \(k\). The matrix $\mathbf{Z'}$ is the embedding matrix for the cell-type network, incorporating readily precomputed elements such as ${\mathbf{X}^{I}}$, ${\mathbf{A'}^{I}_{(1)}\mathbf{X}^{1}}$, and ${\mathbf{A'}^{I}_{(k)}\mathbf{X}^{I}}$. This procedure mirrors the approach adopted in the Voronoi network, with the notable distinction that the stochastic edge sampling technique (yielding $\mathbf{A'}^{I}$) is specifically applied to the cell-type network.

\subsection{Voronoi-Cell-type Attention}
Equipped with two embedding matrices, $\mathbf{Z}$ and $\mathbf{Z'}$, derived from the Voronoi network and the Cell-type network respectively, we now apply Voronoi-Cell-type Attention. This mechanism is designed to autonomously discern the significance of each network's contribution towards relevant downstream tasks, such as binary classification or hazard modeling. Focusing on a specific cell's embedding vector, $\ell$, in image $I$, the attention mechanism operates as follows:

\begin{equation}
\label{eqn:attn_1}
\tilde{\mathbf{z}}_\ell^{I} = \alpha_{\ell,\text{Voronoi}}\mathbf{z}_{\ell}^{I} + \alpha_{\ell,\text{Cell-type}}\mathbf{z'}_{\ell}^{I}
\end{equation}

\noindent where $\tilde{\mathbf{z}}_\ell^{I} \in \mathbb{R}^{D}$ is the resulting embedding vector that encapsulates the significance of each network. Here, attention coefficients are calculated as $\alpha_{\ell,\text{Voronoi}} = \frac{\exp(\hat{a})}{\exp(\hat{a})+\exp(\hat{a'})}$ and $\alpha_{\ell,\text{Cell-type}} = \frac{\exp(\hat{a'})}{\exp(\hat{a})+\exp(\hat{a'})}$, where $\hat{a} = \text{LeakyReLU}(\mathbf{a}^{\top}\mathbf{z}_{\ell}^{I})$ and $\hat{a'} = \text{LeakyReLU}(\mathbf{a}^{\top}\mathbf{z'}_{\ell}^{I})$ indicate attention scores, derived using a learnable vector $\mathbf{a} \in \mathbb{R}^{D}$ and LeakyReLU activation with a negative slope of 0.3. This method discerns the relative importance of the Voronoi (emphasizing coordinate information of neighboring cells or localized tumor microenvironments) and cell-type networks (highlighting specific cell-type populations and related information) for predictions. For binary classification task, the final training loss for~\proposed~integrates both networks' knowledge, calculated via $\mathcal{L_{\text{ce}}} = \sum_{l \in I_{\small{tr}}} CE(\text{Pool}(\mathbf{P}^{l}), \mathcal{Y}^{l})$, where \( CE(\cdot,\cdot) \) represents cross-entropy loss between the pooled predictions \( \mathbf{P} \in \mathbb{R}^{|\mathcal{V}^{l}|\times C} \) and their labels. A 3-layer MLP serves as the prediction head, mapping embeddings to class predictions, with dimensions reflecting nodes \( |\mathcal{V}^{l}| \) and classes \( C \). For the hazard modeling task, Cox partial likelihood substitutes the cross-entropy loss to refine the Stochastic Gradient Descent loss calculation~\cite{cox}.

\section{Experiments}

\noindent{{{\textbf{Experimental Settings. }}} Given that the mIF image benchmark dataset is not widely publicized, we chose to utilize the most recently available datasets from primary human cancer resections, following the precedent set by SPACE-GM~\cite{space-gm}. The data originates from three distinct institutions: the University of Pittsburgh Medical Center (UPMC), Stanford University (Stanford), and the Dana-Farber Cancer Institute (DFCI). This compilation includes three different 40-plex CODEX datasets, totaling 658 sample images. These samples encompass 139 patients diagnosed with head-and-neck cancer (HNC) and 110 patients with colorectal cancer (CRC), resulting in the datasets being categorized as UPMC-HNC, Stanford-CRC, and DFCI-HNC. In the UPMC-HNC dataset, we split the 7 coverslips into 4 for training, 1 for validation, and 2 for testing, and perform binary classification and hazard modeling tasks for patient-level phenotyping. For the Stanford-CRC dataset, given 4 coverslips, we allocate 2 for training, 1 for validation, and 1 for testing, proceeding with the same tasks. To ensure robust predictions, we randomly generated 3 folds for both UPMC-HNC and Stanford-CRC, each with different training, validation, and test coverslips. In the DFCI-HNC dataset, we perform a generalization task using all UPMC-HNC samples for training and evaluate on unseen DFCI-HNC images. We used biomarker expression and cell size as features for both the Voronoi and cell-type networks, while cell type information was used to build the cell-type network. To ensure a fair comparison, we incorporated cell type information as additional features in other baselines. For more detailed experimental settings, including evaluation metrics, baselines, and implementation details, please refer to Appendix~\ref{appendix:C}.

\subsection{Patient-level Phenotype Prediction}

In Tables~\ref{table:table1} and~\ref{table:table2}, we showcase the performance results of phenotype prediction across the UPMC-HNC, Stanford-CNC, and DFCI-HNC datasets. Our observations highlight: \textbf{(1)} Above all, compared to all the baselines with each specific downstream task,~\proposed~consistently outperforms in both Binary Classification and Hazard Modeling, with notable improvements of \underline{21.14}\% and \underline{8.93}\% in Average (BC) compared to the most recent SPACE-GM model and the top-performing GCN model in Table~\ref{table:table2}. \textbf{(2)} Notably,~\proposed~also excels in the Generalization task, with improvements of \underline{8.47}\% over the most recent and top-performing SPACE-GM model. This success is attributed to our unique multiplex network approach, particularly the effective use of a cell-type network in the Generalization task. This network leverages the commonality of cell compositions across domains in mIF images (e.g., B cells and tumor cells), thereby establishing the uniqueness of our work. \textbf{(3)} While multiplex network models like HDMI show potential in the Recurrence task in Binary Classification (see Table~\ref{table:table1}) due to their incorporation of high-order mutual information, the lack of a cell-type network prevents the multiplex network framework from fully benefiting, especially in the mIF image domain. \textbf{(4)} GNNs that manage heterophilous environments, such as FAGCN, encounter difficulties in analyzing mIF images. This underscores that when analyzing mIF images, it is crucial not only to manage the heterophilous environment but also to simultaneously capture geometric information, which aligns with the design principles of~\proposed.

\begin{table}[!t]
\caption{Average performance on test folds for patient-level phenotype prediction using the UPMC-HNC dataset for Binary Classification and Hazard Modeling, and the DFCI-HNC dataset for Generalization. For the Generalization, the model is trained on UPMC-HNC dataset and evaluated on DFCI-HNC dataset. AUC-ROC is applied for Binary Classification and Generalization, while c-index is utilized for Hazard Modeling.}
\centering
\resizebox{1\linewidth}{!}{
\begin{tabular}{@{}l|ccc|c|c|c@{}}
\toprule
           & \multicolumn{4}{c|}{\textbf{Binary Classification (BC)}}        & \textbf{Hazard Modeling} & \textbf{Generalization} \\ \midrule
           & Primary Outcome      & Recurrence           & HPV infection         & \cellcolor[gray]{0.90}\textbf{Average (BC)}   & Survival Length   & Primary Outcome                                     \\ \midrule
GCN\cite{gcn}       & 0.687\scriptsize{$\pm{0.052}$}                & 0.751\scriptsize{$\pm{0.035}$}                & 0.748\scriptsize{$\pm{0.064}$} & 0.729                & 0.693\scriptsize{$\pm{0.045}$}                                             & 0.480   \\
GAT\cite{gat}        & 0.690\scriptsize{$\pm{0.072}$}                 & 0.745\scriptsize{$\pm{0.009}$}                & 0.761\scriptsize{$\pm{0.072}$} & 0.732                 & 0.705\scriptsize{$\pm{0.066}$}                                           & 0.531      \\
GraphSAGE\cite{graphsage}  & 0.708\scriptsize{$\pm{0.063}$}                & 0.743\scriptsize{$\pm{0.034}$}                 & 0.767\scriptsize{$\pm{0.085}$} & 0.739                 & 0.708\scriptsize{$\pm{0.061}$}                                      & 0.430          \\
SIGN\cite{sign}       & 0.715\scriptsize{$\pm{0.030}$}                 & 0.729\scriptsize{$\pm{0.030}$}                & 0.808\scriptsize{$\pm{0.035}$} & 0.751                  & 0.701\scriptsize{$\pm{0.040}$}                                  & 0.467              \\
ClusterGCN\cite{clustergcn} & 0.714\scriptsize{$\pm{0.042}$}                & 0.678\scriptsize{$\pm{0.044}$}                & 0.800\scriptsize{$\pm{0.023}$} & 0.731                 & 0.707\scriptsize{$\pm{0.016}$}                                        & 0.676        \\
FAGCN\cite{fagcn}      & 0.712\scriptsize{$\pm{0.024}$}                & 0.785\scriptsize{$\pm{0.054}$}                & 0.773\scriptsize{$\pm{0.031}$} & 0.757                 & 0.562\scriptsize{$\pm{0.056}$}                                           & 0.524     \\
HDMI\cite{hdmi}       & 0.705\scriptsize{$\pm{0.050}$}                & 0.807\scriptsize{$\pm{0.055}$}                & 0.797\scriptsize{$\pm{0.062}$} & 0.770                 & 0.702\scriptsize{$\pm{0.041}$}                                           & 0.585     \\
SPACE-GM\cite{space-gm}   & 0.716\scriptsize{$\pm{0.059}$}                & 0.767\scriptsize{$\pm{0.036}$}                & 0.778\scriptsize{$\pm{0.052}$} & 0.754                 & 0.600\scriptsize{$\pm{0.129}$}                                          & 0.685      \\ \midrule
\rowcolor[gray]{0.90} \proposed       & \multicolumn{1}{c}{\textbf{0.737}\scriptsize{$\pm{0.060}$}} & \multicolumn{1}{c}{\textbf{0.832}\scriptsize{$\pm{0.065}$}} & \multicolumn{1}{c|}{\textbf{0.813}\scriptsize{$\pm{0.067}$}} & \textbf{0.794} & \textbf{0.728}\scriptsize{$\pm{0.044}$} & \textbf{0.743}                                                \\ \bottomrule

\end{tabular}
}
\label{table:table1}
\end{table}

\begin{table}[!t]
\caption{Average performance on test folds for patient-level phenotype prediction using the Stanford-CNC dataset for Binary Classification and Hazard Modeling. AUC-ROC is applied for Binary Classification, while the c-index is utilized for Hazard Modeling.}
\centering
\resizebox{1\linewidth}{!}{
\begin{tabular}{@{}l|cc|c|cc|c@{}}
\toprule
           & \multicolumn{3}{c|}{\textbf{Binary Classification (BC)}} & \multicolumn{3}{c}{\textbf{Hazard Modeling (HM)}}  \\ \midrule
           & Primary Outcome & Recurrence & \cellcolor[gray]{0.90}\textbf{Average (BC)} & Survival Length & Recurrence Interval & \cellcolor[gray]{0.90}\textbf{Average (HM)} \\ \cmidrule{2-7}
GCN\cite{gcn}        & 0.579\scriptsize{$\pm{0.036}$} & 0.630\scriptsize{$\pm{0.098}$} & 0.605 & 0.571\scriptsize{$\pm{0.079}$} & 0.568\scriptsize{$\pm{0.087}$} & 0.570 \\
GAT\cite{gat}        & 0.539\scriptsize{$\pm{0.022}$} & 0.524\scriptsize{$\pm{0.043}$} & 0.532 & 0.603\scriptsize{$\pm{0.012}$} & 0.567\scriptsize{$\pm{0.037}$} & 0.585 \\
GraphSAGE\cite{graphsage}  & 0.540\scriptsize{$\pm{0.110}$} & 0.572\scriptsize{$\pm{0.093}$} & 0.556 & 0.578\scriptsize{$\pm{0.068}$} & 0.507\scriptsize{$\pm{0.055}$} & 0.543 \\
SIGN\cite{sign}       & 0.522\scriptsize{$\pm{0.008}$} & 0.466\scriptsize{$\pm{0.076}$} & 0.494 & 0.541\scriptsize{$\pm{0.096}$} & 0.553\scriptsize{$\pm{0.082}$} & 0.547 \\
ClusterGCN\cite{clustergcn} & 0.545\scriptsize{$\pm{0.059}$} & 0.531\scriptsize{$\pm{0.060}$} & 0.538 & 0.473\scriptsize{$\pm{0.136}$} & 0.484\scriptsize{$\pm{0.047}$} & 0.479 \\
FAGCN\cite{fagcn}      & 0.584\scriptsize{$\pm{0.092}$} & 0.498\scriptsize{$\pm{0.081}$} & 0.541 & 0.567\scriptsize{$\pm{0.037}$} & 0.502\scriptsize{$\pm{0.021}$} & 0.535 \\
HDMI\cite{hdmi}       & 0.499\scriptsize{$\pm{0.026}$} & 0.483\scriptsize{$\pm{0.028}$} & 0.491 & 0.565\scriptsize{$\pm{0.076}$} & 0.572\scriptsize{$\pm{0.033}$} & 0.569 \\
SPACE-GM\cite{space-gm}   & 0.563\scriptsize{$\pm{0.035}$} & 0.524\scriptsize{$\pm{0.041}$} & 0.544 & 0.492\scriptsize{$\pm{0.059}$} & 0.577\scriptsize{$\pm{0.043}$} & 0.535 \\ \midrule
\rowcolor[gray]{0.90} \proposed      & \textbf{0.658}\scriptsize{$\pm{0.030}$} & \textbf{0.660}\scriptsize{$\pm{0.047}$} & \textbf{0.659} & \textbf{0.631}\scriptsize{$\pm{0.048}$} & \textbf{0.597}\scriptsize{$\pm{0.083}$} & \textbf{0.614} \\ \bottomrule
\end{tabular}
}
\label{table:table2}
\end{table}

\subsection{In-depth analysis of~\proposed}

\begin{figure*}[!h]
    \centering
    \includegraphics[width=1.0\columnwidth]{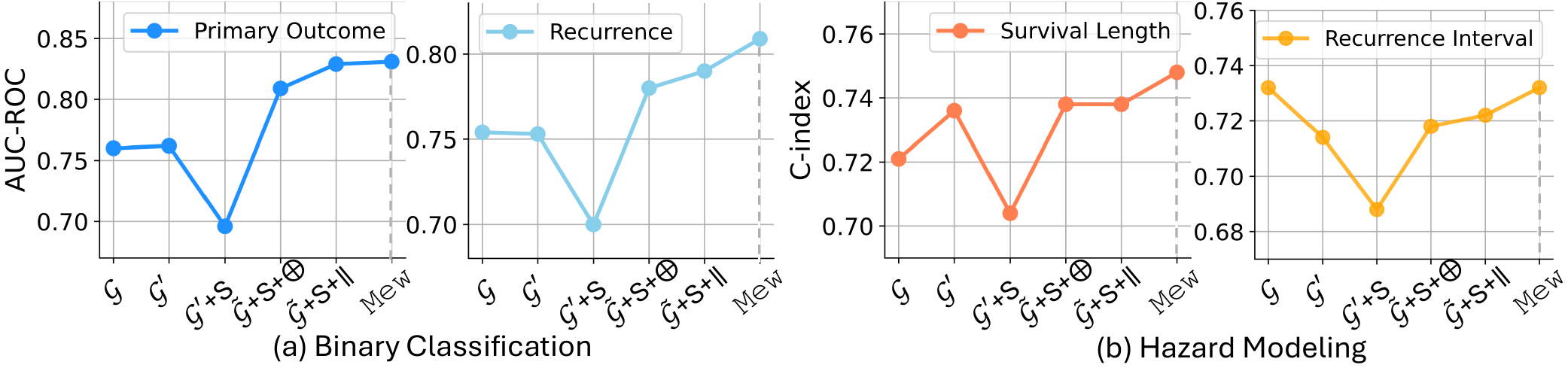}
    \caption{Ablation study of~\proposed~on the Stanford-CNC dataset for (a) Binary Classification (b) Hazard Modeling. Here, $\mathcal{G}$, $\mathcal{G'}$, and $\tilde{\mathcal{G}}$ denote the Voronoi network, Cell-type network, and Multiplex network, respectively.~`\textsf{S}' indicates the Stochastic Edge Sampling technique, while `$\oplus$' and `$||$' represent addition and concatenation operations, respectively. Applying the attention mechanism to the multiplex network completes~\proposed.}
    \label{fig:ablation}
\end{figure*}

\noindent{{{\textbf{Ablation Studies. }}} In Figure~\ref{fig:ablation}, we unveil key insights: \textbf{(1)} A synergistic benefit is observed when combining both the Voronoi network ($\mathcal{G}$) and the Cell-type network ($\mathcal{G'}$) to generate a Multiplex network ($\tilde{\mathcal{G}}$), rather than relying solely one of them. \textbf{(2)} The Stochastic Edge Sampling technique ($\textsf{S}$) fully leverages its benefits within the Multiplex network framework, as its application solely to the Cell-type network omits information from the Voronoi network. \textbf{(3)} Given a Multiplex network, the optimal strategy for information fusion transcends mere addition or concatenation; the integration of the Voronoi-Cell-type Attention emerges as the most effective approach. \textbf{(4)} In Figure~\ref{fig:ablation} (b), within the `Recurrence Interval' plot, although the performance of~\proposed~and the Voronoi network ($\mathcal{G}$) seem comparable, it is crucial to note that a multi-task learning setting is employed for each patient-level phenotyping task, such as Binary Classification or Hazard Modeling. This distinction becomes evident in the `Survival Length' task, where~\proposed~exhibits superior performance compared to using the Voronoi network alone, thereby underscoring the effectiveness of our proposed framework.

\begin{figure*}[!ht]
    \centering
    \includegraphics[width=1\columnwidth]{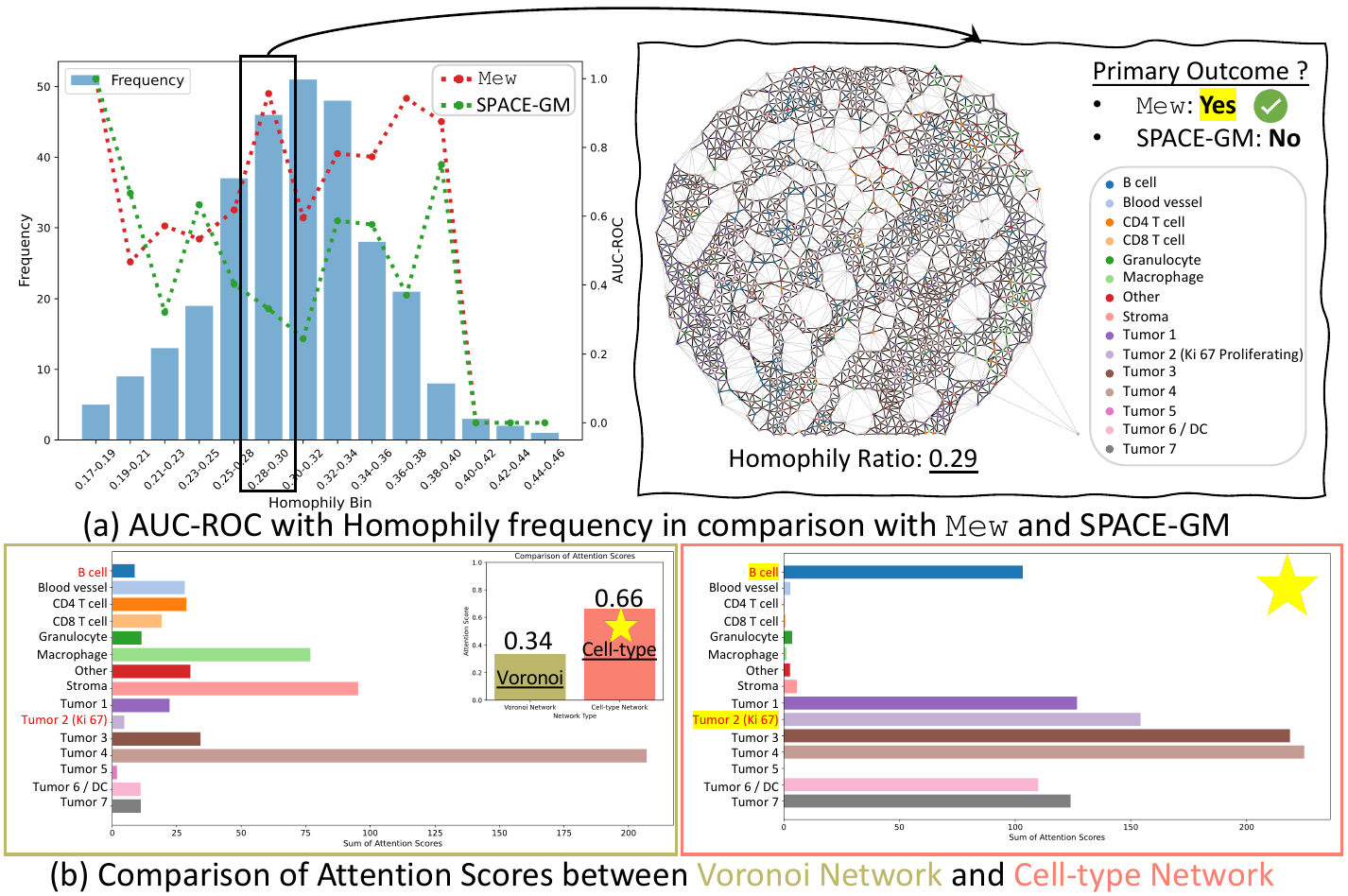}
    \caption{In-depth Analysis of~\proposed~on the Stanford-CRC Dataset: (a) Graphs are categorized based on their homophily ratios, and the corresponding AUC-ROC scores are compared with those from SPACE-GM. Notably,~\proposed~demonstrates a marked performance improvement at lower homophily rates, with a representative graph exemplifying this trend. (b) A comparison of attention scores within the Voronoi Network and Cell-type Network reveals distinct functions of each network. In the highlighted graph, the emphasis on the cell-type network—characterized by significant tumor expression—facilitates accurate classification as a primary outcome case.}
    \label{fig:fig4}
\end{figure*}

\noindent{{{\textbf{How Cell-type network contributes? }}} 
Here, we explore the effectiveness of~\proposed, with a focus on the Cell-type network's contribution to heterophilous scenarios. Figure~\ref{fig:fig4} (a) showcases a comparison between the performance of~\proposed~and SPACE-GM. The histogram plot reveals that most images fall within low homophily rates (e.g., 0.25-0.34), where the performance disparity between~\proposed~and SPACE-GM becomes evident. We delved deeper into a scenario where the performance disparity was most pronounced, notably in the 0.28 to 0.30 bin where~\proposed~accurately identified the phenotype and included a representative cellular graph example from that bin. As illustrated by Figure~\ref{fig:fig4} (b), our analysis of the distribution of attention scores across layers highlighted the Cell-type network's critical role, with an average attention score of 0.66, in contrast to the Voronoi network's 0.34. Importantly, the Cell-type network highlighted the significance of various tumor cells, like Tumor 2 (Ki67 Proliferating) — a marker for proliferation rates — and highlighted B cells, signaling an immune response to the tumor. This level of detail and insight was unattainable with the Voronoi network alone, which primarily captures local geometric heterogeneity. Overall, these findings emphasize the Cell-type network's capacity to improve phenotype prediction, providing valuable interpretability in heterophilous settings.

\subsection{Scalability and Generalizability of~\proposed}
\label{sec:scalability}

\noindent{{{\textbf{Scalability. }}} This section highlights the criticality of scalability in the preprocessing, training, and evaluation phases of the graph model. Table~\ref{table:table3} demonstrates how \proposed~adeptly addresses scalability challenges. This is largely attributed to the implementation of scalable GNNs, which significantly reduce preprocessing time compared to SPACE-GM. SPACE-GM traditionally saves 3-hop neighbors as chunks prior to training, resulting in considerable time complexity and memory usage, often exceeding an hour. While this approach may expedite the training process, the overwhelming preprocessing costs diminish overall efficiency.~\proposed, on the other hand, offers two primary advantages: \textbf{(1)} It broadens the node perspective by incorporating the entire graph directly ($|\bar{\mathcal{V}}|$=117,974 versus SPACE-GM's $|\bar{\mathcal{V}}|$=1,071), avoiding the need for manual sampling and storing of 3-hop local graphs as chunks, thereby reducing preprocessing time significantly. \textbf{(2)} It achieves remarkably low evaluation times due to the upfront preprocessing efforts (0.53s to evaluate 53 graphs versus SPACE-GM's 60.28s). Once the precomputation of \(\mathbf{A}\mathbf{X}\) is completed, the bulk of computational work shifts to the forward pass of the weight parameters and the precomputed ones. As a result,~\proposed~stands out as an exceptionally efficient and feasible model for real-world phenotype prediction, where swift decision-making is crucial.

\begin{table*}[!t]
\centering
\caption{Scalability analysis of the UPMC-HNC dataset. $|\bar{\mathcal{V}}|$ denotes the average number of nodes handled in the graphs. `chunk-save' saves 3-hop subgraphs in memory during preprocessing and utilizes them during training and evaluation, whereas `on-the-fly' samples subgraphs instantly. During training, time is measured up to 100 iterations.}

\resizebox{1.0\linewidth}{!}{
\begin{tabular}{@{}l|ccc@{}}
\toprule
                & SPACE-GM (chunk-save)    & SPACE-GM (on-the-fly)      & \proposed                         \\ \midrule
Preprocessing ($|\mathcal{G}|=308$) & 5345.12s                  & -                          & 75.60s                       \\
Training ($|\mathcal{G}|=16$)  & 7.03s ($|\bar{\mathcal{V}}|=1,071$) & 20.45s ($|\bar{\mathcal{V}}|=1,071$) & 36.01s ($|\bar{\mathcal{V}}|=117,974$) \\
Evaluation ($|\mathcal{G}|=53$)     & 60.28s                    & 156.18s                    & 0.53s                        \\ \bottomrule
\end{tabular}
}
\label{table:table3}
\end{table*}

\noindent{{{\textbf{Generalizability. }}} To achieve greater generalizability in real-world scenarios where cell-type annotations may be unavailable, a straightforward remedy is to utilize and harmonize with recent advancements in cell-type annotation methods~\cite{ct_annotation1, ct_annotation2} to supplement annotations, enabling the construction of cell-type networks for running~\proposed. When annotations are partially available, pseudo-labeling methods such as Label Propagation~\cite{lp_main} can be used. In extreme cases with no annotations, K-Means clustering~\cite{kmeans} can provide a cell type index. We verify applicability in such cases using the Broad Bioimage Benchmark Collection and demonstrate generalizability to other domains such as Whole Slide Images in Appendix~\ref{appendix:D}.

\section{Limitation}
Despite the effectiveness and efficiency of~\proposed~in processing mIF images, relying exclusively on image modality in real-world scenarios may lead to suboptimal outcomes for patient-level phenotype prediction. Intriguingly, our future work will aim to incorporate additional modalities, such as genomic and clinical data, to enhance the final prediction accuracy. In this context, we anticipate that the current multiplex network can be further expanded by adding layers specific to each modality, provided common properties, such as cell indices, are available. This approach promises to significantly enrich the model's interpretability and predictive power by leveraging the synergies between different types of data.

\section{Conclusion}
In this paper, we address two critical challenges inherent in applying spatial graph-based ML to mIF images:~\ding{172}~\underline{Cellular Heterogeneity} and~\ding{173}~\underline{Scalability} of mIF cellular graphs. Recognizing the heterogeneous nature of cellular data, we propose the generation of a \textbf{Multiplex Network} by incorporating an additional Cell-type network, naturally complementing the inductive biases of GNNs. Furthermore, to ensure practical applicability in real-world scenarios, we have developed a \textbf{scalable Graph Neural Networks} equipped with a novel stochastic edge sampling technique. This architecture is further enhanced by a \textbf{Voronoi-Cell-type Attention} which assesses the significance of each network. Rigorous testing on real-world patient datasets has consistently highlighted the robustness and superior efficacy of our proposed method,~\proposed, showcasing its potential to pioneer a new and promising direction for advancing mIF image analysis.

\section*{Acknowledgements}
We thank the reviewers for their helpful comments. We also thank Emanuele Rossi for the insightful discussion regarding the usage of SIGN. This work was partially supported by the National Research Foundation of Korea(NRF) grant funded by the Korea government(MSIT) (RS-2024-00335098).

%
%
\bibliographystyle{splncs04}
\bibliography{main}

\begin{thebibliography}{10}
\providecommand{\url}[1]{\texttt{#1}}
\providecommand{\urlprefix}{URL }
\providecommand{\doi}[1]{https://doi.org/#1}

\bibitem{prml}
Bishop, C.M.: Pattern recognition and machine learning. Springer google schola  \textbf{2},  5--43 (2006)

\bibitem{louvain}
Blondel, V.D., Guillaume, J.L., Lambiotte, R., Lefebvre, E.: Fast unfolding of communities in large networks. Journal of statistical mechanics: theory and experiment  \textbf{2008}(10),  P10008 (2008)

\bibitem{fagcn}
Bo, D., Wang, X., Shi, C., Shen, H.: Beyond low-frequency information in graph convolutional networks. In: Proceedings of the AAAI Conference on Artificial Intelligence. vol.~35, pp. 3950--3957 (2021)

\bibitem{gatv2}
Brody, S., Alon, U., Yahav, E.: How attentive are graph attention networks? arXiv preprint arXiv:2105.14491  (2021)

\bibitem{bbbc_v1}
Caie, P.D., Walls, R.E., Ingleston-Orme, A., Daya, S., Houslay, T., Eagle, R., Roberts, M.E., Carragher, N.O.: High-content phenotypic profiling of drug response signatures across distinct cancer cells. Molecular cancer therapeutics  \textbf{9}(6),  1913--1926 (2010)

\bibitem{ct_annotation1}
Chen, J., Xu, H., Tao, W., Chen, Z., Zhao, Y., Han, J.D.J.: Transformer for one stop interpretable cell type annotation. Nature Communications  \textbf{14}(1), ~223 (2023)

\bibitem{clustergcn}
Chiang, W.L., Liu, X., Si, S., Li, Y., Bengio, S., Hsieh, C.J.: Cluster-gcn: An efficient algorithm for training deep and large graph convolutional networks. In: Proceedings of the 25th ACM SIGKDD international conference on knowledge discovery \& data mining. pp. 257--266 (2019)

\bibitem{3d_unet}
{\c{C}}i{\c{c}}ek, {\"O}., Abdulkadir, A., Lienkamp, S.S., Brox, T., Ronneberger, O.: 3d u-net: learning dense volumetric segmentation from sparse annotation. In: Medical Image Computing and Computer-Assisted Intervention--MICCAI 2016: 19th International Conference, Athens, Greece, October 17-21, 2016, Proceedings, Part II 19. pp. 424--432. Springer (2016)

\bibitem{chebynet}
Defferrard, M., Bresson, X., Vandergheynst, P.: Convolutional neural networks on graphs with fast localized spectral filtering. Advances in neural information processing systems  \textbf{29},  3844--3852 (2016)

\bibitem{macro_tumor}
DeNardo, D.G., Brennan, D.J., Rexhepaj, E., Ruffell, B., Shiao, S.L., Madden, S.F., Gallagher, W.M., Wadhwani, N., Keil, S.D., Junaid, S.A., et~al.: Leukocyte complexity predicts breast cancer survival and functionally regulates response to chemotherapy. Cancer discovery  \textbf{1}(1),  54--67 (2011)

\bibitem{traidtional_image1}
Egmont-Petersen, M., de~Ridder, D., Handels, H.: Image processing with neural networks—a review. Pattern recognition  \textbf{35}(10),  2279--2301 (2002)

\bibitem{mIF3}
Fassler, D.J., Abousamra, S., Gupta, R., Chen, C., Zhao, M., Paredes, D., Batool, S.A., Knudsen, B.S., Escobar-Hoyos, L., Shroyer, K.R., et~al.: Deep learning-based image analysis methods for brightfield-acquired multiplex immunohistochemistry images. Diagnostic pathology  \textbf{15}(1),  1--11 (2020)

\bibitem{sign}
Frasca, F., Rossi, E., Eynard, D., Chamberlain, B., Bronstein, M., Monti, F.: Sign: Scalable inception graph neural networks. arXiv preprint arXiv:2004.11198  (2020)

\bibitem{gan_appication_1}
Frid-Adar, M., Diamant, I., Klang, E., Amitai, M., Goldberger, J., Greenspan, H.: Gan-based synthetic medical image augmentation for increased cnn performance in liver lesion classification. Neurocomputing  \textbf{321},  321--331 (2018)

\bibitem{neighbor1}
Giraldo, N.A., Sanchez-Salas, R., Peske, J.D., Vano, Y., Becht, E., Petitprez, F., Validire, P., Ingels, A., Cathelineau, X., Fridman, W.H., et~al.: The clinical role of the tme in solid cancer. British journal of cancer  \textbf{120}(1),  45--53 (2019)

\bibitem{traidtional_image0}
Gonzales, R.C., Wintz, P.: Digital image processing. Addison-Wesley Longman Publishing Co., Inc. (1987)

\bibitem{deeplearning_ian}
Goodfellow, I., Bengio, Y., Courville, A.: Deep learning. MIT press (2016)

\bibitem{gan}
Goodfellow, I., Pouget-Abadie, J., Mirza, M., Xu, B., Warde-Farley, D., Ozair, S., Courville, A., Bengio, Y.: Generative adversarial networks. Communications of the ACM  \textbf{63}(11),  139--144 (2020)

\bibitem{deepcell}
Greenwald, N.F., Miller, G., Moen, E., Kong, A., Kagel, A., Dougherty, T., Fullaway, C.C., McIntosh, B.J., Leow, K.X., Schwartz, M.S., et~al.: Whole-cell segmentation of tissue images with human-level performance using large-scale data annotation and deep learning. Nature biotechnology  \textbf{40}(4),  555--565 (2022)

\bibitem{traidtional_image3}
Gurcan, M.N., Boucheron, L.E., Can, A., Madabhushi, A., Rajpoot, N.M., Yener, B.: Histopathological image analysis: A review. IEEE reviews in biomedical engineering  \textbf{2},  147--171 (2009)

\bibitem{neighbor2}
Gut, G., Herrmann, M.D., Pelkmans, L.: Multiplexed protein maps link subcellular organization to cellular states. Science  \textbf{361}(6401),  eaar7042 (2018)

\bibitem{graphsage}
Hamilton, W.L., Ying, R., Leskovec, J.: Inductive representation learning on large graphs. In: Proceedings of the 31st International Conference on Neural Information Processing Systems. pp. 1025--1035 (2017)

\bibitem{hu20212020}
Hu, X., Chang, M.C., Chen, Y., Sridhar, R., Hu, Z., Xue, Y., Wu, Z., Pi, P., Shen, J., Tan, J., et~al.: The 2020 low-power computer vision challenge. In: 2021 IEEE 3rd International Conference on Artificial Intelligence Circuits and Systems (AICAS). pp.~1--4. IEEE (2021)

\bibitem{hu2021e2vts}
Hu, Z., Pi, P., Wu, Z., Xue, Y., Shen, J., Tan, J., Lian, X., Wang, Z., Liu, J.: E2vts: energy-efficient video text spotting from unmanned aerial vehicles. In: Proceedings of the IEEE/CVF conference on computer vision and pattern recognition. pp. 905--913 (2021)

\bibitem{image_to_image}
Isola, P., Zhu, J.Y., Zhou, T., Efros, A.A.: Image-to-image translation with conditional adversarial networks. In: Proceedings of the IEEE conference on computer vision and pattern recognition. pp. 1125--1134 (2017)

\bibitem{hdmi}
Jing, B., Park, C., Tong, H.: Hdmi: High-order deep multiplex infomax. In: Proceedings of the Web Conference 2021. pp. 2414--2424 (2021)

\bibitem{gcn}
Kipf, T.N., Welling, M.: Semi-supervised classification with graph convolutional networks. arXiv preprint arXiv:1609.02907  (2016)

\bibitem{kmeans}
Krishna, K., Murty, M.N.: Genetic k-means algorithm. IEEE Transactions on Systems, Man, and Cybernetics, Part B (Cybernetics)  \textbf{29}(3),  433--439 (1999)

\bibitem{cox}
Kvamme, H., Borgan, {\O}., Scheel, I.: Time-to-event prediction with neural networks and cox regression. arXiv preprint arXiv:1907.00825  (2019)

\bibitem{cora}
Lin, F., Cohen, W.W.: Semi-supervised classification of network data using very few labels. In: 2010 International Conference on Advances in Social Networks Analysis and Mining. pp. 192--199. IEEE (2010)

\bibitem{mIF2}
Lin, J.R., Fallahi-Sichani, M., Sorger, P.K.: Highly multiplexed imaging of single cells using a high-throughput cyclic immunofluorescence method. Nature communications  \textbf{6}(1), ~8390 (2015)

\bibitem{bbbc_og}
Ljosa, V., Sokolnicki, K.L., Carpenter, A.E.: Annotated high-throughput microscopy image sets for validation. Nature methods  \textbf{9}(7),  637--637 (2012)

\bibitem{macro_tcell}
Mantovani, A., Sica, A., Sozzani, S., Allavena, P., Vecchi, A., Locati, M.: The chemokine system in diverse forms of macrophage activation and polarization. Trends in immunology  \textbf{25}(12),  677--686 (2004)

\bibitem{mIF4}
Maric, D., Jahanipour, J., Li, X.R., Singh, A., Mobiny, A., Van~Nguyen, H., Sedlock, A., Grama, K., Roysam, B.: Whole-brain tissue mapping toolkit using large-scale highly multiplexed immunofluorescence imaging and deep neural networks. Nature communications  \textbf{12}(1), ~1550 (2021)

\bibitem{mIF5}
Maric, D., Jahanipour, J., Li, X.R., Singh, A., Mobiny, A., Van~Nguyen, H., Sedlock, A., Grama, K., Roysam, B.: Whole-brain tissue mapping toolkit using large-scale highly multiplexed immunofluorescence imaging and deep neural networks. Nature communications  \textbf{12}(1), ~1550 (2021)

\bibitem{birds}
McPherson, M., Smith-Lovin, L., Cook, J.M.: Birds of a feather: Homophily in social networks. Annual review of sociology  \textbf{27}(1),  415--444 (2001)

\bibitem{apache}
Meng, X., Bradley, J., Yavuz, B., Sparks, E., Venkataraman, S., Liu, D., Freeman, J., Tsai, D., Amde, M., Owen, S., et~al.: Mllib: Machine learning in apache spark. The journal of machine learning research  \textbf{17}(1),  1235--1241 (2016)

\bibitem{cgan}
Mirza, M., Osindero, S.: Conditional generative adversarial nets. arXiv preprint arXiv:1411.1784  (2014)

\bibitem{gan_appication_2}
Osokin, A., Chessel, A., Carazo~Salas, R.E., Vaggi, F.: Gans for biological image synthesis. In: Proceedings of the IEEE International Conference on Computer Vision. pp. 2233--2242 (2017)

\bibitem{dmgi}
Park, C., Kim, D., Han, J., Yu, H.: Unsupervised attributed multiplex network embedding. In: Proceedings of the AAAI Conference on Artificial Intelligence. vol.~34, pp. 5371--5378 (2020)

\bibitem{ct_annotation2}
Pasquini, G., Arias, J.E.R., Sch{\"a}fer, P., Busskamp, V.: Automated methods for cell type annotation on scrna-seq data. Computational and Structural Biotechnology Journal  \textbf{19},  961--969 (2021)

\bibitem{pytorch}
Paszke, A., Gross, S., Chintala, S., Chanan, G., Yang, E., DeVito, Z., Lin, Z., Desmaison, A., Antiga, L., Lerer, A.: Automatic differentiation in pytorch  (2017)

\bibitem{macro_homo2}
Qian, B.Z., Pollard, J.W.: Macrophage diversity enhances tumor progression and metastasis. Cell  \textbf{141}(1),  39--51 (2010)

\bibitem{dcgan}
Radford, A., Metz, L., Chintala, S.: Unsupervised representation learning with deep convolutional generative adversarial networks. arXiv preprint arXiv:1511.06434  (2015)

\bibitem{spatial2}
Rodriques, S.G., Stickels, R.R., Goeva, A., Martin, C.A., Murray, E., Vanderburg, C.R., Welch, J., Chen, L.M., Chen, F., Macosko, E.Z.: Slide-seq: A scalable technology for measuring genome-wide expression at high spatial resolution. Science  \textbf{363}(6434),  1463--1467 (2019)

\bibitem{unet}
Ronneberger, O., Fischer, P., Brox, T.: U-net: Convolutional networks for biomedical image segmentation. In: Medical Image Computing and Computer-Assisted Intervention--MICCAI 2015: 18th International Conference, Munich, Germany, October 5-9, 2015, Proceedings, Part III 18. pp. 234--241. Springer (2015)

\bibitem{citation}
Sen, P., Namata, G., Bilgic, M., Getoor, L., Galligher, B., Eliassi-Rad, T.: Collective classification in network data. AI magazine  \textbf{29}(3),  93--93 (2008)

\bibitem{traidtional_image2}
Shen, D., Wu, G., Suk, H.I.: Deep learning in medical image analysis. Annual review of biomedical engineering  \textbf{19},  221--248 (2017)

\bibitem{shen2021learning}
Shen, J., Chen, X., Heaton, H., Chen, T., Liu, J., Yin, W., Wang, Z.: Learning a minimax optimizer: A pilot study. In: International Conference on Learning Representations (2021), \url{https://openreview.net/forum?id=nkIDwI6oO4_}

\bibitem{shen2021umec}
Shen, J., Wang, H., Gui, S., Tan, J., Wang, Z., Liu, J.: {\{}UMEC{\}}: Unified model and embedding compression for efficient recommendation systems. In: International Conference on Learning Representations (2021), \url{https://openreview.net/forum?id=BM---bH_RSh}

\bibitem{macro_homo1}
Sica, A., Mantovani, A., et~al.: Macrophage plasticity and polarization: in vivo veritas. The Journal of clinical investigation  \textbf{122}(3),  787--795 (2012)

\bibitem{coauthor}
Sinha, A., Shen, Z., Song, Y., Ma, H., Eide, D., Hsu, B.J., Wang, K.: An overview of microsoft academic service (mas) and applications. In: Proceedings of the 24th international conference on world wide web. pp. 243--246 (2015)

\bibitem{morpho_image}
Soille, P., et~al.: Morphological image analysis: principles and applications, vol.~2. Springer (1999)

\bibitem{mIF1}
Sood, A., Miller, A.M., Brogi, E., Sui, Y., Armenia, J., McDonough, E., Santamaria-Pang, A., Carlin, S., Stamper, A., Campos, C., et~al.: Multiplexed immunofluorescence delineates proteomic cancer cell states associated with metabolism. JCI insight  \textbf{1}(6) (2016)

\bibitem{spatial1}
St{\aa}hl, P.L., Salm{\'e}n, F., Vickovic, S., Lundmark, A., Navarro, J.F., Magnusson, J., Giacomello, S., Asp, M., Westholm, J.O., Huss, M., et~al.: Visualization and analysis of gene expression in tissue sections by spatial transcriptomics. Science  \textbf{353}(6294),  78--82 (2016)

\bibitem{efficient_net}
Tan, M., Le, Q.: Efficientnet: Rethinking model scaling for convolutional neural networks. In: International conference on machine learning. pp. 6105--6114. PMLR (2019)

\bibitem{transformer}
Vaswani, A., Shazeer, N., Parmar, N., Uszkoreit, J., Jones, L., Gomez, A.N., Kaiser, {\L}., Polosukhin, I.: Attention is all you need. Advances in neural information processing systems  \textbf{30} (2017)

\bibitem{gat}
Veli{\v{c}}kovi{\'c}, P., Cucurull, G., Casanova, A., Romero, A., Lio, P., Bengio, Y.: Graph attention networks. arXiv preprint arXiv:1710.10903  (2017)

\bibitem{space-gm}
Wu, Z., Trevino, A.E., Wu, E., Swanson, K., Kim, H.J., D’Angio, H.B., Preska, R., Charville, G.W., Dalerba, P.D., Egloff, A.M., et~al.: Graph deep learning for the characterization of tumour microenvironments from spatial protein profiles in tissue specimens. Nature Biomedical Engineering  \textbf{6}(12),  1435--1448 (2022)

\bibitem{gin}
Xu, K., Hu, W., Leskovec, J., Jegelka, S.: How powerful are graph neural networks? arXiv preprint arXiv:1810.00826  (2018)

\bibitem{yu2022unified}
Yu, S., Chen, T., Shen, J., Yuan, H., Tan, J., Yang, S., Liu, J., Wang, Z.: Unified visual transformer compression. arXiv preprint arXiv:2203.08243  (2022)

\bibitem{lp_main}
Zhu, X.: Semi-supervised learning with graphs. Carnegie Mellon University (2005)

\end{thebibliography}

\newpage
\appendix

\section{Additional Data Statistics}\label{appendix:A}

Besides the UPMC-HNC used in the main manuscript, Figure~\ref{fig:fig_stanford} and Figure~\ref{fig:fig_dfci} show the statistics of the Stanford-CRC and DFCI-HNC datasets, respectively.

\begin{figure}[!th]
    \centering
    \includegraphics[width=0.7\columnwidth]{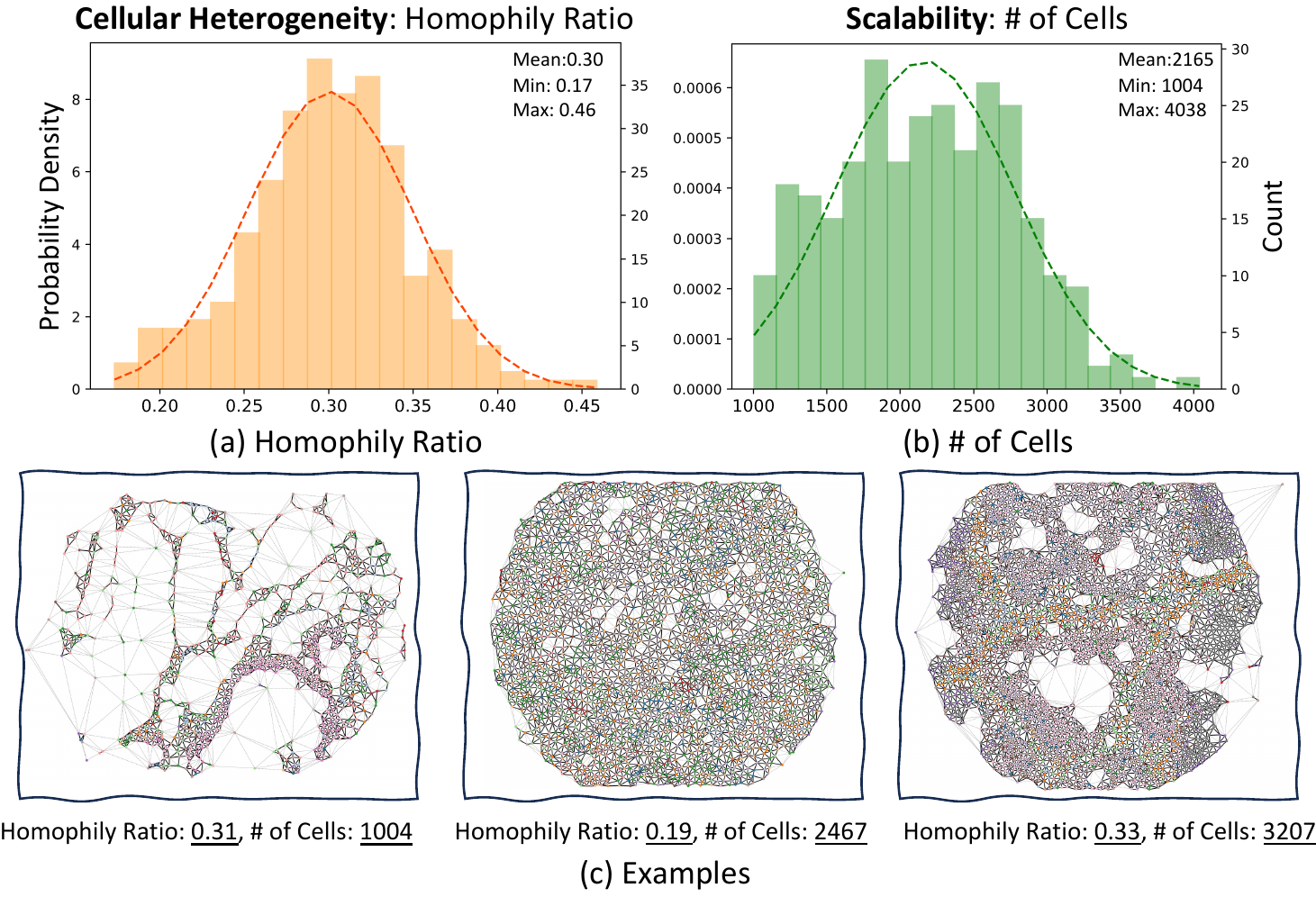}
    \vspace{-3mm}
    \caption{Statistics for the 292 samples in the Stanford-CRC dataset, detailing (a) Cellular Heterogeneity: Distribution of homophily ratios, and (b) Scalability: Distribution of cell counts per graph. The mean, minimum, and maximum values are included for each distribution (\textit{e.g.}, at upper corners). (c) Example of generated graphs.}
    \label{fig:fig_stanford}
    \vspace{-3mm}
\end{figure}

\begin{figure}[!h]
    \centering
    \includegraphics[width=0.7\columnwidth]{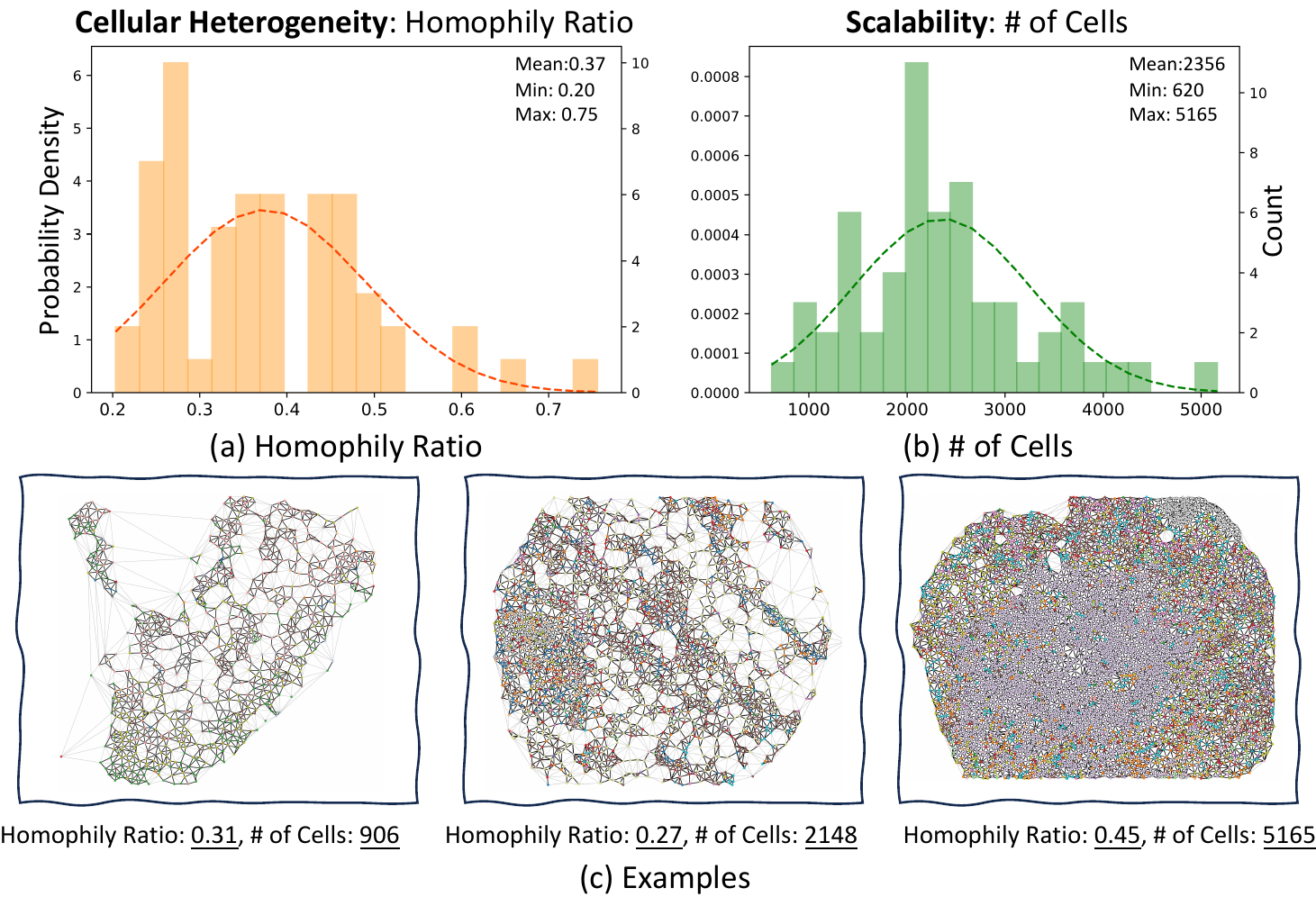}
    \vspace{-2mm}
    \caption{Statistics for the 58 samples in the DFCI-HNC dataset, detailing (a) Cellular Heterogeneity: Distribution of homophily ratios, and (b) Scalability: Distribution of cell counts per graph. The mean, minimum, and maximum values are included for each distribution (\textit{e.g.}, at upper corners). (c) Example of generated graphs.}
    \label{fig:fig_dfci}
    \vspace{-3mm}
\end{figure}

\section{Overall Algorithm of~\proposed}\label{appendix:B}
Here, we present the comprehensive training algorithm of ~\proposed~ as outlined in Algorithm~\ref{alg:miim}. Utilizing the Voronoi graph as a base, we further construct the Cell-type network by connecting nodes of common cell types (Line 2). The algorithm involves precomputing the matrix multiplication of the adjacency matrix and feature matrix (Lines 5 and 6). During the training phase, we leverage the precomputed output to implement Voronoi-Cell-type attention (Line 12). The process then involves computing the cross-entropy loss and updating weight parameters until convergence is achieved.

\begin{algorithm}
\small
\caption{Pseudocode of the proposed algorithm}\label{alg:miim}
\hspace*{\algorithmicindent} \textbf{Input:} mpIF image $I$ with its corresponding phenotype $\mathcal{Y}^{I}$ \\
\hspace*{\algorithmicindent} \textbf{Output:} Pooled Prediction Matrix $\mathbf{P}^{I}$
\begin{algorithmic}[1]

\State $\mathcal{G}^{I} \gets (\mathcal{V}^{I}, \mathcal{E}^{I})$ \Comment{Voronoi network via preprocessing image $I$}
\State $\mathcal{E'}^{I} \gets \Call{\textcolor{blue}{Cell-type Edge Generater}}{\mathcal{C}^{I}, \mathcal{V}^{I}}$
\State $\mathcal{G'}^{I} \gets (\mathcal{V}^{I}, \mathcal{E'}^{I})$ \Comment{Cell-type network}
\State $\tilde{\mathcal{G}}^{I} \gets (\mathcal{G}^{I}, \mathcal{G'}^{I})$
\Comment{\textit{Multiplex Network: Voronoi + Cell-type}}
\State $\textcolor{orange}{L} \gets \Call{\textcolor{blue}{Precomputer}}{\mathbf{A}^{I}, \mathbf{X}^{I}, K, \text{stochastic=}False}$
\State $\textcolor{orange}{L^{'}} \gets \Call{\textcolor{blue}{Precomputer}}{\mathbf{A'}^{I}, \mathbf{X}^{I}, K, \text{stochastic=}True}$

\While{not converged}
\State $\mathbf{H}^{I} = \sigma([\textcolor{orange}{L[0]}\mathbf{W}_{(0)},\textcolor{orange}{L[1]}\mathbf{W}_{(1)}\ldots, \textcolor{orange}{L[K]}\mathbf{W}_{(K)})$
\State $\mathbf{Z}^{I} = \xi({\mathbf{H}^{I}\mathbf{W}_{z}})$
\State $\mathbf{H'}^{I} = \sigma([\textcolor{orange}{L'[0]}\mathbf{W}_{(0)},\textcolor{orange}{L'[1]}\mathbf{W}_{(1)}\ldots, \textcolor{orange}{L'[K]}\mathbf{W}_{(K)})$
\State $\mathbf{Z'}^{I} = \xi({\mathbf{H'}^{I}\mathbf{W}_{z}})$

\State $\tilde{\mathbf{Z}}^{I} \gets \Call{\textcolor{blue}{Voronoi-Cell-type Attention}}{\mathbf{Z}^{I}, \mathbf{Z'}^{I}}$
\State $\mathbf{P}^{l} \gets MLP(\tilde{\mathbf{Z}}^{l})$
\State $\mathcal{L_{\text{ce}}} \gets CE(\text{Pool}(\mathbf{P}^{I}), \mathcal{Y}^{I})$

\EndWhile

\Function{\textcolor{blue}{Cell-type Edge Generater}}{$\mathcal{C}^{I}, \mathcal{V}^{I}$}
\State {$\mathcal{E'}^{I} = \{(\mathcal{V'}^{I}_{i}, \mathcal{V'}^{I}_{j}) \mid \mathcal{C}_i^{I} = \mathcal{C}_j^{I}, \forall i, j \in \{1, ..., |\mathcal{V'}^{I}|\}, i \neq j \}$}

\State \textbf{return} $\mathcal{E'}^{I}$
\EndFunction

\Function{\textcolor{blue}{Precomputer}}{$\mathbf{A}^{I}, \mathbf{X}^{I}, K, \text{stochastic=}False$}
            \State Initialize empty list $L = []$
            \State Append $\mathbf{X}^{I}$ to $L$
            \For{$k \gets 1 \text{ to } K+1$}
            \If{\text{stochastic}}
            \State $\mathbf{A}_{ij}^{I} = 
            \begin{cases}
            1 & \text{if}\;\; \text{Bernoulli}(\mathbf{\hat{P}}_{ij}^{I}) = 1 \\
            0 & \text{Otherwise}
            \end{cases}
            $
            \EndIf
            \State $\mathbf{X}^{I} \gets \mathbf{A}^{I}\mathbf{X}^{I}$
            \State Append $\mathbf{X}^{I}$ to $L$
            \EndFor            
            \State \textbf{return} $L$
            \EndFunction

\Function{\textcolor{blue}{Voronoi-Cell-type Attention}}{$\mathbf{Z}^{I}, \mathbf{Z'}^{I}$}
\For{$\ell \gets 0 \text{ to } |\mathcal{V}^{I}|$}
\State{$\alpha_{\ell,\text{Voronoi}} = \frac{\exp({\text{LeakyReLU}(\mathbf{a}^{\top}\mathbf{z}_{\ell}^{I})})}{\exp({\text{LeakyReLU}(\mathbf{a}^{\top}\mathbf{z}_{\ell}^{I})}) + \exp({\text{LeakyReLU}(\mathbf{a}^{\top}\mathbf{z'}_{\ell}^{I})})}$} 
\State{$\alpha_{\ell,\text{Cell-type}} = \frac{\exp({\text{LeakyReLU}(\mathbf{a}^{\top}\mathbf{z'}_{\ell}^{I})})}{\exp({\text{LeakyReLU}(\mathbf{a}^{\top}\mathbf{z}_{\ell}^{I})}) + \exp({\text{LeakyReLU}(\mathbf{a}^{\top}\mathbf{z'}_{\ell}^{I})})}
$}
\State{$\tilde{\mathbf{z}}_\ell^{I} = \alpha_{\ell,\text{Voronoi}}\mathbf{z}_{\ell}^{I} + \alpha_{\ell,\text{Cell-type}}\mathbf{z'}_{\ell}^{I}$}
\EndFor
\State \textbf{return} $\tilde{\mathbf{Z}}^{I}$
\EndFunction
\end{algorithmic}
\end{algorithm}

\section{Detailed Experimental Settings}\label{appendix:C}

\paragraph{Evaluation metrics. }
We evaluate binary classification tasks using the receiver operating characteristic curve (AUC-ROC) and hazard modeling tasks using the C-index. Particularly, the C-index measures the proportion of all possible observation pairs for which the predicted values match the actual survival order. We use the following formula to calculate the C-index:
\begin{equation}\label{eq:cindex}
\text{C-index} = \frac{1}{n(n - 1)} \sum_{i=1}^n \sum_{j=1}^n I(T_i < T_j)(1 - c_{j}),
\end{equation}
which involves the following variables: $n$ (sample size), $\mathrm{T}_{i}$ and $\mathrm{T}_{j}$ (representing the survival time of the $i$-th and $j$-th patients, respectively).
The $\mathrm{I}(\cdot)$ denotes the indicator function, which evaluates to $1$ if its argument is true and $0$ otherwise. 
Additionally, ${c}_{j}$ denotes the correct censorship status.

\paragraph{Baselines. }
Given our focus on effective and efficient graph learning in mIF imaging, we benchmark our model against various state-of-the-art graph learning models. Our comparison spans a range of GNNs from the machine learning literature, including foundational models like GCN~\cite{gcn}, GAT~\cite{gat}, and GraphSAGE~\cite{graphsage}; scalable solutions such as SIGN~\cite{sign} and ClusterGCN~\cite{clustergcn}; a heterophilous model: FAGCN~\cite{fagcn}; a multiplex network model: HDMI~\cite{hdmi}; and a recent model from the biology literature: SPACE-GM~\cite{space-gm}. The deatils of each baseline can be described as below:

\begin{itemize}
\item \textbf{GCN}~\cite{gcn}: learn node representations by aggregating features from their neighbors using a form of convolution adapted for graph data. This approach efficiently captures the local graph structure and node features.

\item \textbf{GAT}~\cite{gat}: generates node embeddings by sampling and aggregating features from a node's local neighborhood. It can efficiently generate embeddings for large graphs by leveraging node feature information.

\item \textbf{GraphSAGE}~\cite{graphsage}: introduce attention mechanisms to graph neural networks, allowing for more expressive representations by weighting the significance of each neighbor's features during aggregation differently.

\item \textbf{SIGN}~\cite{sign}: extends the idea of inception modules from convolutional networks to graph neural networks, enabling it to capture different ranges of neighborhood information without the need for costly recursive neighborhood expansions.

\item \textbf{ClusterGCN}~\cite{clustergcn}: improves the scalability of graph convolutional networks by partitioning the graph into clusters and then performing GCN within each cluster, significantly reducing computational and memory requirements.

\item \textbf{FAGCN}~\cite{fagcn}: addresses the limitations of existing GNNs by adaptively integrating both low-frequency and high-frequency signals during message passing. 

\item \textbf{HDMI}~\cite{hdmi}: enhances node embedding on multiplex networks through a novel framework. It leverages high-order mutual information, combining intrinsic and extrinsic signals, and employs an attention-based fusion module for integrating embeddings from different network layers.

\item \textbf{SPACE-GM}~\cite{space-gm}: is a novel graph deep learning framework that characterizes tumor microenvironments by leveraging spatial protein profiles in tissue specimens. It models tumor microenvironments as local subgraphs, capturing distinctive cellular interactions associated with different clinical outcomes.

\end{itemize}

\paragraph{Impelmentation Details. }
To ensure a fair comparison of each model, we trained all baselines using the Adam optimizer.
We adopt $1000$ training epochs for all datasets and models.
To set hyper-parameters, we explored hyper-parameters within the following rages: learning rate $\{0.0001, 0.001, 0.01, 0.1\}$, sub-graph size $\{1, 2, 3, 4\}$, hidden dimension $\{64, 128, 256, 512\}$, training batch size$\{16, 32, 64\}$, and the dropout ratio $\{0.0, 0.25, 0.5\}$. It is noteworthy that aside from the hyperparameter controlling the shared weight matrix across the Voronoi and Cell-type networks, our proposed model,~\proposed, requires no model-specific hyperparameters for meticulous adjustment. The multiplex network generation, stochastic edge sampling, and Voronoi-Cell-type attention module inherently leverage and autonomously determine their significance.

\paragraph{Best Hyperparamters. }
The best hyperparameters for~\proposed~are provided in Table~\ref{tab:hyperparamsetting}.

\begin{table}[!h]
\centering
\caption{The best hyperparameter setup for~\proposed. `BC' and `HM' denote Binary Classification and Hazard Modeling, respectively.}
\resizebox{0.8\linewidth}{!}{
\begin{tabular}{@{}l|cc|cc|c@{}}
\toprule
           & \multicolumn{2}{c|}{UPMC-HNC} & \multicolumn{2}{c|}{Stanford-CRC} & \multicolumn{1}{c}{DFCI-HNC} \\ \midrule
           & BC & HM & BC & HM & BC     \\ 
           \midrule
    Shared & True & True & False & False & False \\
    Learning rate & $0.001$ & $0.0001$ & $0.0001$ & $0.0001$ & $0.001$ \\
    \# of layers & $3$ & $4$ & $4$ & $4$ & $2$ \\
    Hidden dimension & $512$ & $512$ & $512$ & $128$ & $128$ \\
    Batch size & $32$ & $16$ & $16$ & $32$ & $16$ \\
    Dropout ratio & $0.0$ & $0.0$ & $0.0$ & $0.5$ & $0.25$ \\
\bottomrule
\end{tabular}
\label{tab:hyperparamsetting}
}
\end{table}

\section{Detailed Generalizability of~\proposed.}\label{appendix:D}

\noindent \textbf{When cell-type annotation is unavailable. }
In practice, creating a cell-type network within a multiplex framework may face hurdles due to the need for cell-type annotation, often requiring manual effort. To circumvent this for patient-level phenotype prediction, we propose utilizing a \underline{clustering index} as a surrogate for manually annotated cell-types, allowing for the construction of cell-type networks by connecting cells with identical clustering indexes. This method significantly reduces the dependency on manual input while adhering to the multiplex network framework's principles. Although it limits certain post-hoc analyses, such as marker gene identification, the clustering index approach is advantageous for our main objective of phenotype prediction. For further exploration of this strategy in practical scenarios, we detail its application on the benchmark image set BBBC021v1~\cite{bbbc_v1} from the Broad Bioimage Benchmark Collection~\cite{bbbc_og} in the subsequent paragraph.

\noindent \textbf{Benchmark result on BBBC021. }
To broaden the application of our method to benchmark datasets, we chose to implement it on the BBBC021 dataset. This dataset is ideal for evaluating image-based profiling methods' ability to predict drug mechanisms of action. It comprises images from a p53-normal breast cancer model (MCF-7), annotated with standard morphological markers and featuring a diverse array of both targeted and cytotoxic agents. These agents induce a wide range of phenotypes, enabling an in-depth analysis of their effects. However, unlike the dataset used in the main manuscript, BBBC021 only offers image data and its corresponding Mechanisms-of-Action (MoA), serving as the target label for each image, but lacks information on cell coordinates, biomarker expression, and cell-type annotation. To address these limitations, we first \textbf{(1)} apply DeepCell~\cite{deepcell} to segment the cells, obtaining cell coordinate and biomarker expression data. Then, in situations where cell-type annotation is unavailable, \textbf{(2)} we employ K-Means~\cite{kmeans} clustering to process the biomarker expression data, using the resulting cluster index as a proxy for cell-type. From the 10 weeks of available image datasets, Weeks 8 (292 images), 9 (408 images), and 10 (240 images) were specifically selected as training, validation, and test sets, respectively. Figure~\ref{fig:bbbc} depicts the comprehensive procedure and the accuracy achieved in MoA prediction.

\begin{figure}[!ht]
    \centering
    \includegraphics[width=1\columnwidth]{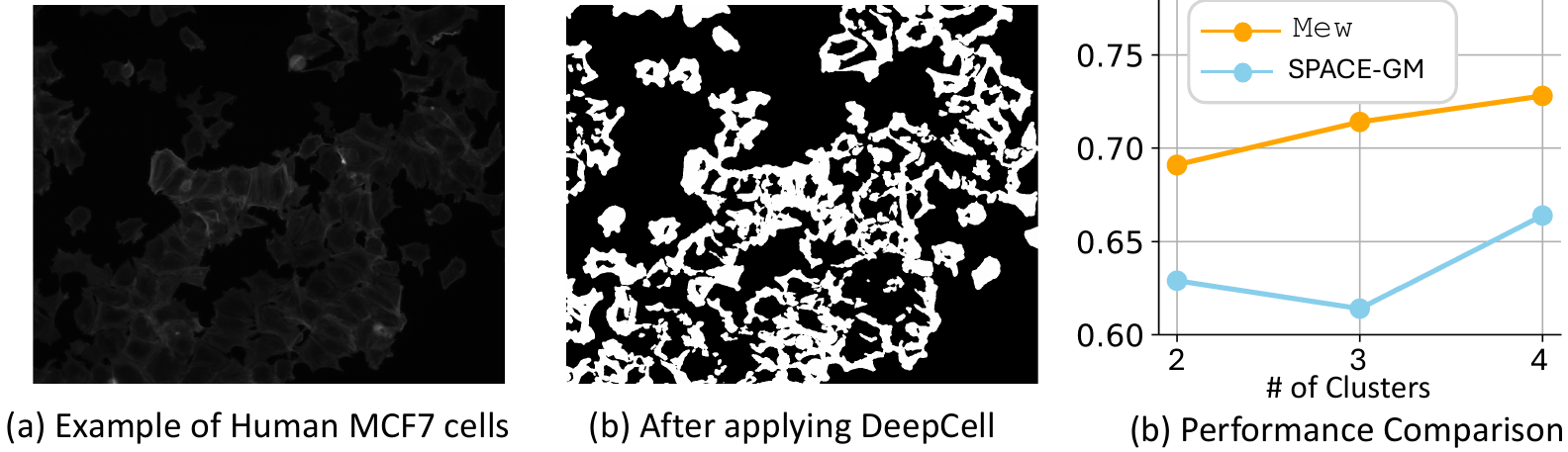}
    \vspace{-4mm}
    \caption{Preprocessing of the BBBC021 dataset and performance comparison: (a) An example image of Human MCF7 breast cancer cells. (b) After processing with DeepCell, we obtain a segmented image that provides cell coordinates and biomarker information. (c) Utilizing the K-Means algorithm, we substitute cell-type information with a cluster index. By varying the number of clusters in the K-Means algorithm, we evaluate the effectiveness of the multiplex network approach,~\proposed, compared to SPACE-GM.}
    \label{fig:bbbc}
    \vspace{-3mm}
\end{figure}

\noindent \textbf{Applicability to WSIs. } Adapting multiplexed immunofluorescence (mIF) image analysis to the Whole Slide Images (WSIs) domain requires meticulous consideration of several key aspects. The substantial increase in resolution and size of WSIs, compared to standard mIF images, necessitates robust data management and processing strategies for effective data handling. In this context, the employment of scalable Graph Neural Networks presents a promising solution, as equipped within~\proposed. Furthermore, given the wide morphological diversity observed in WSIs, precise annotation and efficient selection of regions of interest are imperative. In scenarios where annotation is challenging, the use of a clustering index offers a viable alternative as mentioned above. Additionally, it is essential to optimize the imaging system's multiplexing capacity to facilitate the simultaneous detection of multiple biomarkers. This shift also calls for significant computational resources to process and analyze the data effectively. Integrating mIF data from WSIs with clinical information and leveraging advanced image analysis algorithms are crucial steps towards extracting meaningful biological insights, underscoring the importance of a thorough data analysis approach.

\end{document}